\PassOptionsToPackage{table}{xcolor}
\documentclass[sigconf]{acmart}
\usepackage{algorithm}
\usepackage{multirow}
\usepackage{booktabs}
\usepackage{subcaption}
\usepackage{comment}
\usepackage{url}
\usepackage{ulem}
\useunder{\uline}{\ul}{}
\usepackage{enumitem}
\usepackage[misc]{ifsym}

\AtBeginDocument{%
  }

\setcopyright{acmlicensed}
\copyrightyear{2018}
\acmYear{2018}
\acmDOI{XXXXXXX.XXXXXXX}
\acmConference[Conference acronym ’XX]{}{June 03–05, 2018,}{Woodstock, NY}
\acmISBN{978-1-4503-XXXX-X/18/06}

\begin{document}

\title{Lightweight Channel-wise Dynamic Fusion Model: Non-stationary Time Series Forecasting via Entropy Analysis}

\author{Tianyu Jia}
\affiliation{%
  \institution{Tianjin University}
  \city{Tianjin}
  \country{China}}
\email{jiatianyu@tju.edu.cn}

\author{Zongxia Xie \Letter}
\affiliation{%
  \institution{Tianjin University}
  \city{Tianjin}
  \country{China}}
\email{caddiexie@hotmail.com}

\author{Yanru Sun}
\affiliation{%
  \institution{Tianjin University}
  \city{Tianjin}
  \country{China}}
\email{yanrusun@tju.edu.cn}

\author{Dilfira Kudrat}
\affiliation{%
  \institution{Tianjin University}
  \city{Tianjin}
  \country{China}}
\email{dilfira\_49@tju.edu.cn}

\author{Qinghua Hu}
\affiliation{%
  \institution{Tianjin University}
  \city{Tianjin}
  \country{China}}
\email{huqinghua@tju.edu.cn}

\renewcommand{\shortauthors}{Tianyu Jia et al.}

\begin{abstract}
Non-stationarity is an intrinsic property of real-world time series and plays a crucial role in time series forecasting. 
Previous studies primarily adopt instance normalization to attenuate the non-stationarity of original series for better predictability.
However, instance normalization that directly removes the inherent non-stationarity can lead to three issues: (1) disrupting global temporal dependencies, (2) ignoring channel-specific differences, and (3) producing over-smoothed predictions.
To address these issues, we theoretically demonstrate that variance can be a valid and interpretable proxy for quantifying non-stationarity of time series.
Based on the analysis, we propose a novel lightweight \textit{C}hannel-wise \textit{D}ynamic \textit{F}usion \textit{M}odel (\textit{CDFM}), which selectively and dynamically recovers intrinsic non-stationarity of the original series, while keeping the predictability of normalized series.
First, we design a Dual-Predictor Module, which involves two branches: a Time Stationary Predictor for capturing stable patterns and a Time Non-stationary Predictor for modeling global dynamics patterns.
Second, we propose a Fusion Weight Learner to dynamically characterize the intrinsic non-stationary information across different samples based on variance. 
Finally, we introduce a Channel Selector to selectively recover non-stationary information from specific channels by evaluating their non-stationarity, similarity, and distribution consistency, enabling the model to capture relevant dynamic features and avoid overfitting.
Comprehensive experiments on seven time series datasets demonstrate the superiority and generalization capabilities of CDFM.
\end{abstract}

\maketitle

\section{Introduction}
Time series forecasting has been increasingly important in various fields, including finance \cite{huang2024generative}, weather prediction \cite{han2021joint, bi2023accurate}, electricity consumption \cite{amber2015electricity} and so on \cite{liu2023cross}. 

Recently, deep learning models have shown significant improvements by effectively capturing sequential dependencies, such as TimesNet \cite{wu2023timesnet}, PatchTST \cite{nie2023time}, iTransformer \cite{liu2024itransformer} and others \cite{li2023revisiting, zhou2023one, zeng2023transformers, wang2025mamba, ahamed2024timemachine}. 
However, non-stationarity, characterized by the continuous change of joint distribution over time, affects the prediction accuracy and generalization performance \cite{du2021adarnn}.

To mitigate non-stationarity issues, instance normalization methods like RevIN \cite{kim2021reversible}, Dish-TS \cite{fan2023dish} and SAN \cite{liu2024adaptive} have been extensively adopted by deep learning models 

as a simple yet effective data pre-processing technique. 
Current approaches primarily focus on removing non-stationary signals of raw time series by subtracting the mean and dividing by the variance, which helps to provide more stable data distribution for deep models \cite{kim2021reversible}.
However, while these normalization methods have improved predictive performance, non-stationarity is an inherent characteristic of real-world time series and can also provide valuable guidance for discovering temporal dependencies \cite{liu2022non, ye2024frequency, ma2024u}. 
Simply removing non-stationary information from the original series can lead to three main issues that reduce the model's predictive capability:

\begin{figure*}[!t]
\centering
\includegraphics[width=0.83\textwidth]{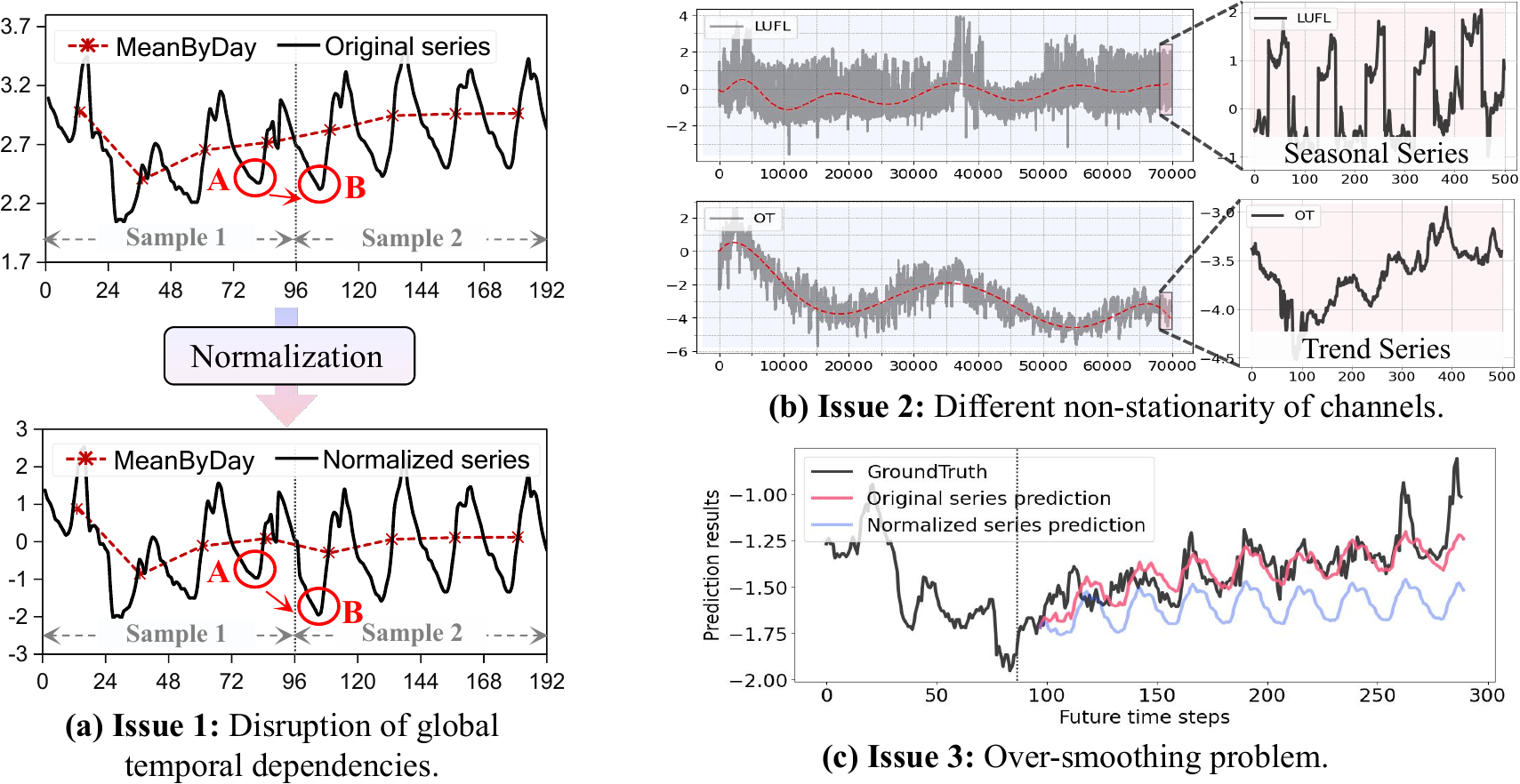} 
\caption{Issues caused by instance normalization. (a) Instance normalization changes the original temporal dependencies. (b) Different channels exhibit unique non-stationarity, such as seasonal and trend channels. (c) The over-smoothing problem caused by instance normalization. These predictions are obtained by DLinear \cite{zeng2023transformers}.}
\label{fig: fig1}
\end{figure*}

Firstly, the normalization methods \textbf{disrupt the global temporal dependencies}, as illustrated in Fig.~\ref{fig: fig1} (a).
The red dashed lines show that an upward trend that is evident before normalization becomes less visible after normalization. 
Furthermore, normalization alters the relative magnitudes among windows, amplifying differences between points \cite{park2024self}, such as point A and point B .
This distortion reduces the model's sensitivity to temporal context and diminishes its ability to capture global temporal dependencies. \cite{park2024self, liu2024TimeBridge}.

Secondly, \textbf{different channels exhibit unique non-stationary properties}, as depicted in Fig.~\ref{fig: fig1} (b). 
However, normalization methods uniformly remove non-stationary information across all channels, which fails to account for the heterogeneity of different features \cite{ye2024frequency}.
For instance, the 'LUFL' channel exhibits a clear seasonal pattern, while the 'OT' channel displays a stronger trend component, making it more susceptible to the effect of normalization \cite{li2023revisiting}. Furthermore, normalization aligns sample distributions and blurs the distinct non-stationary patterns of different channels, resulting in over-smoothed predictions.

Finally, we observe that training on the normalized series leads to the \textbf{over-smoothing problem} \cite{sun2024learning}, causing the predictions to overlook diversity and complexity \cite{liu2022non}, as shown in Fig.~\ref{fig: fig1}(c).
This phenomenon limits the model’s predictive ability and may even lead to substantial deviations from the ground truth, such as overestimating energy consumption on calm days when actual usage is low, and underestimating energy consumption during peak demand periods when actual usage is high \cite{sun2024learning}. More details are discussed in Appendix~\ref{sec: over-smoothing}.

To address the challenges, we propose a lightweight \textbf{C}hannel-wise \textbf{D}ynamic \textbf{F}usion \textbf{M}odel (CDFM) that selectively and dynamically recovers the intrinsic non-stationary characteristics of the original series across specific channels, thus capturing global temporal dependencies and avoiding over-smoothing problem.
Specifically, CDFM consists of the following three key components.
Firstly, we propose a \textbf{Dual-Predictor Module}, which involves two complementary branches: the Time Stationary Predictor adopts instance normalization to mitigate the non-stationarity, enabling accurate predictions for stationary components, while the Time Non-stationary Predictor directly predicts the original series to intuitively capture original distributional shifts and global temporal dependencies.
Secondly, we design a \textbf{Fusion Weight Learner}, which utilizes variance as a measure of non-stationary to dynamically adjust the fusion weight, controlling the recovery ratio of non-stationary information.
Finally, the \textbf{Channel Selector} identifies specific channels by evaluating their non-stationarity, similarity, and distribution consistency and selectively recovers their information, preserving significant non-stationary dynamics while filtering out irrelevant information and avoiding overfitting.
By integrating non-stationary information, CDFM mitigates over-smoothing and ensures the preservation of essential temporal dependencies.

In summary, the contributions of this paper are as follows: 

\begin{itemize}[left=0em]
\item We propose a distribution-free method for measuring the non-stationarity of time series, supported by both theoretical and experimental analyses.
\item We propose CDFM, a lightweight and effective model that dynamically integrates non-stationary information into the forecasting process, improving temporal modeling and addressing over-smoothing challenges.
\item Comprehensive experiments on seven real-world datasets demonstrate the superiority performance of CDFM. Furthermore, our fusion framework demonstrates remarkable generalization capabilities.
\end{itemize}

\section{RELATED WORK}

\subsection{Normalization for Time Series Forecasting}
Existing methods for addressing non-stationarity of time series typically normalize the data to enhance predictability.  
Normalization methods have been extensively explored 
and adopted by deep learning models.
RevIN \cite{kim2021reversible} first introduces a two-stage symmetric reversible normalization method. 
It involves normalizing the input sequences and subsequently denormalizing the model output sequences. Building on this concept, Dish-TS \cite{fan2023dish} addresses intra-sample distributional shift between the historical and horizon windows by separately learning the statistics for normalization and denormalization. SAN \cite{liu2024adaptive} further focuses on intra-sample fine distributional shift, which first splits the sample into non-overlapping subsequence-level patches and applies normalization-and-denormalization to each patch. Based on RevIN \cite{kim2021reversible}, NLinear \cite{zeng2023transformers} proposes a simpler normalization method that subtracts and adds the last value of each sample. SIN \cite{han2024sin} proposes a selective and interpretable normalization that selects statistical data and learns normalization transformations by partial least squares.
Given the limitation of the above methods on temporal domain, FAN \cite{ye2024frequency} proposes a frequency adaptive normalization method.

However, directly applying such normalization to remove non-stationary information from the original series prevents the model from intuitively learning changes in the original data distribution, potentially limiting its capability to capture original temporal dependencies \cite{liu2022non, wang2024considering}. 

\subsection{Non-stationary Models}
Recently, some approaches attempt to leverage non-stationarity within original series for forecasting.
KNF \cite{yu2023koopman} and Koopa \cite{liu2023koopa} leverage modern Koopman theory to predict highly non-stationary time series, providing a systematic approach to uncovering the underlying dynamics of complex systems.
NS-Trans \cite{liu2022non} tackles non-stationary problems by approximating distinguishable attentions learned from original series. 
HTV-Trans \cite{wang2024considering} develops a hierarchical generative module to capture the multi-scale non-stationary information of original series and recovers this non-stationary information into stationary input.
TFPS \cite{sun2024learning} proposes a pattern-aware method that recovers intrinsic non-stationarity in the latent representation and adapts experts to evolving data patterns.

\begin{figure}[t]
    \begin{subfigure}{0.22\textwidth}
        \centerline{
        \includegraphics[width=\linewidth]{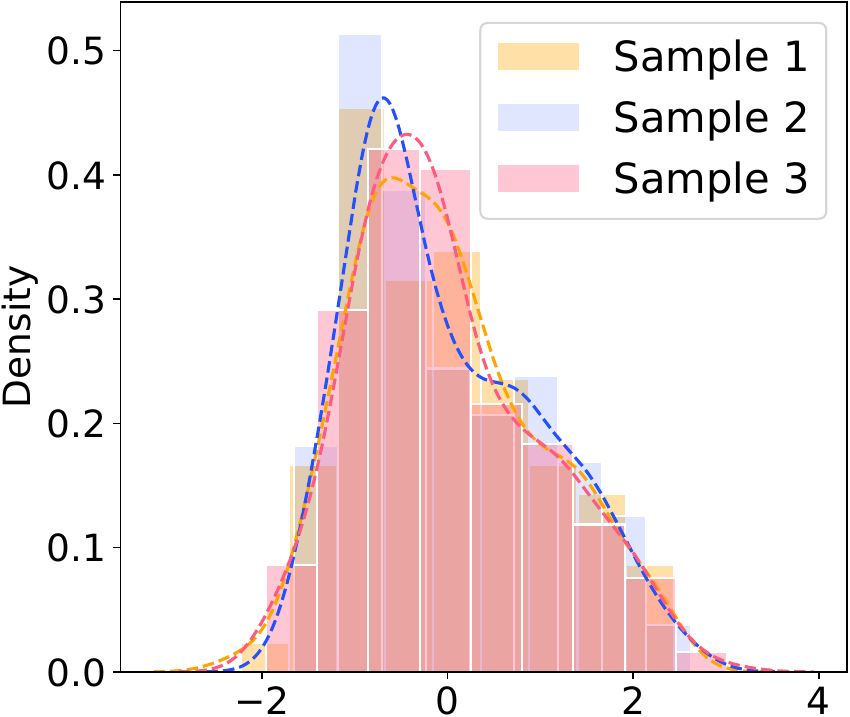}}
        \caption{Normalized distribution}
        \label{fig:fig2-left}
    \end{subfigure}
    \hspace{0.02\textwidth}
    \begin{subfigure}{0.22\textwidth}
        \centerline{
        \includegraphics[width=\linewidth]{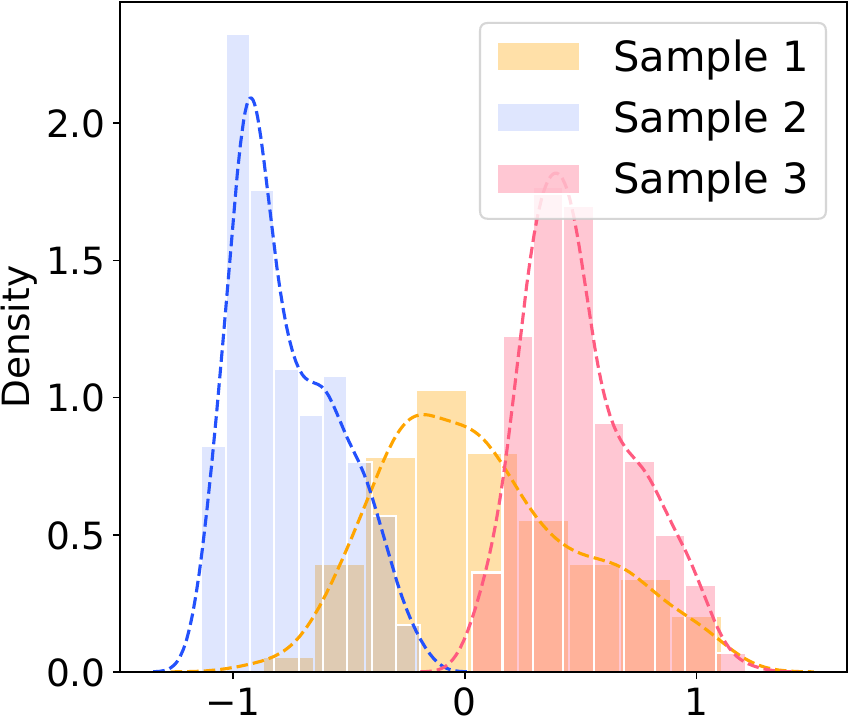}}
        \caption{Original distribution}
        \label{fig:fig2-right}
    \end{subfigure}
    \caption{Normalization mitigates the distribution discrepancy between different samples. }
    \label{fig: fig2}
\end{figure}

However, most existing non-stationary models are limited to the transformer architecture, which has high computational complexity and memory costs \cite{xu2024fits}. In addition, these non-stationary models uniformly restore non-stationary information across all channels without considering the differing properties of them, which can increase the uncertainty and complexity for forecasting and even cause overfitting issues.

Recognizing the limitations of both normalization methods and non-stationary models, we propose a lightweight model architecture that normalizes time series for better predictability while selectively and dynamically recovers non-stationary information of specific channel to capture crucial temporal dependencies. To correctly position CDFM, we also provide a detailed review of the development of time series forecasting methods in Appendix~\ref{sec: deep models for time series forecasting}.

\section{Problem Formulations}
We define the entire time series as \(X= \{ x_1,x_2,...,x_T \} \in \mathbb{R}^{T \times N}\), where \(T\) is the length of time series, \(N\) is the number of channels and \(x_i\) is the observation at the \(i\)-th time point with \(N\) dimensions. We select a sample \(\mathcal{S}_t=(\mathcal{X}_t, \mathcal{Y}_t)\) from \(X\), where \(\mathcal{X}_t=\{x_t,..., x_{t+L-1}\}\) is input sequence with a historical windows length \(L\) and \(\mathcal{Y}_t=\{x_{t+L},..., x_{t+L+H-1}\}\) is the corresponding horizon sequence with a horizon windows length \(H\). The goal of time series forecasting task is to learn a mapping function \(\mathcal{F}\) that maps \(\mathcal{X}_t\) to \(\mathcal{Y}_t\).

\section{Non-stationary Analysis via Entropy}
In this section, we first give the definition of instance normalization.
Then, we conduct a quantitative analysis of non-stationarity in terms of information entropy. Finally, we point out that variance is a effective metric to quantify the degree of non-stationarity.

\begin{figure}[t]
    \centering
    \begin{subfigure}{0.22\textwidth}
        \centering
        \includegraphics[width=\linewidth]{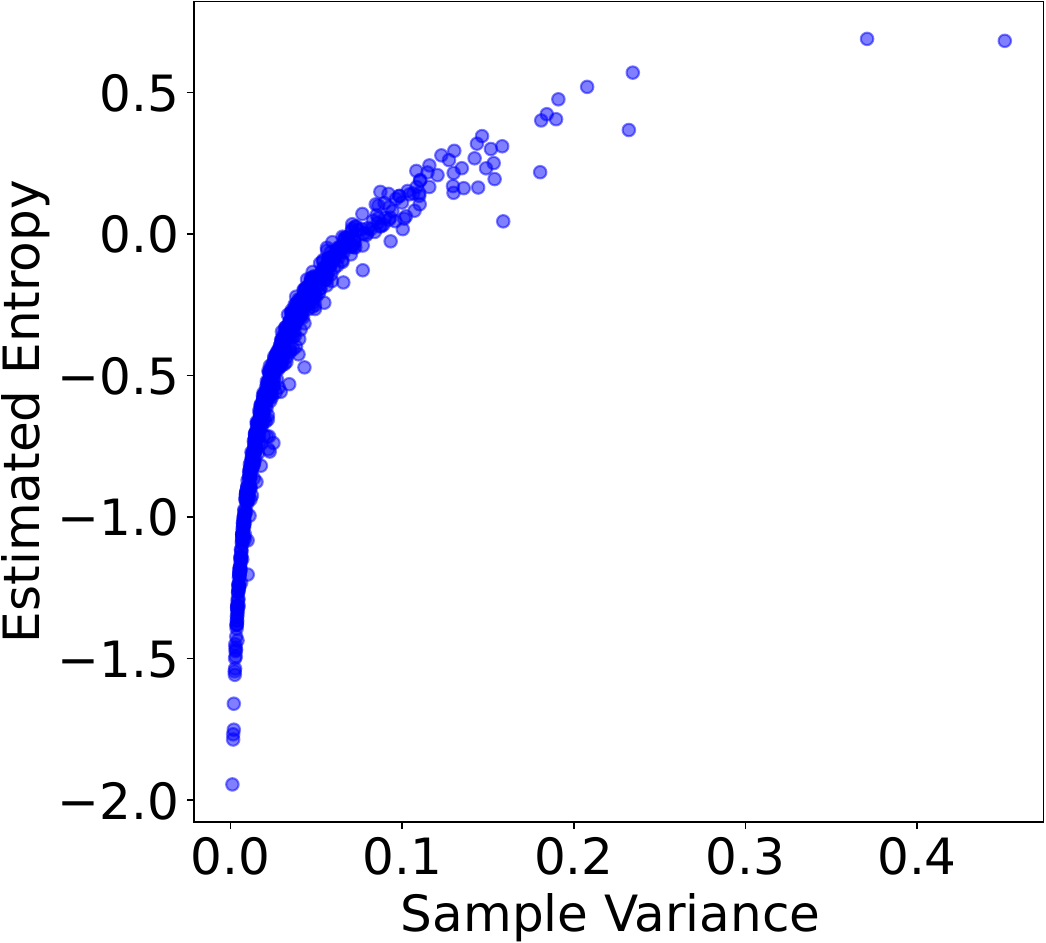}
        \caption{HUFL}
        \label{fig:fig3-left}
    \end{subfigure}
    \hspace{0.02\textwidth} 
    \begin{subfigure}{0.22\textwidth}
        \centering
        \includegraphics[width=\linewidth]{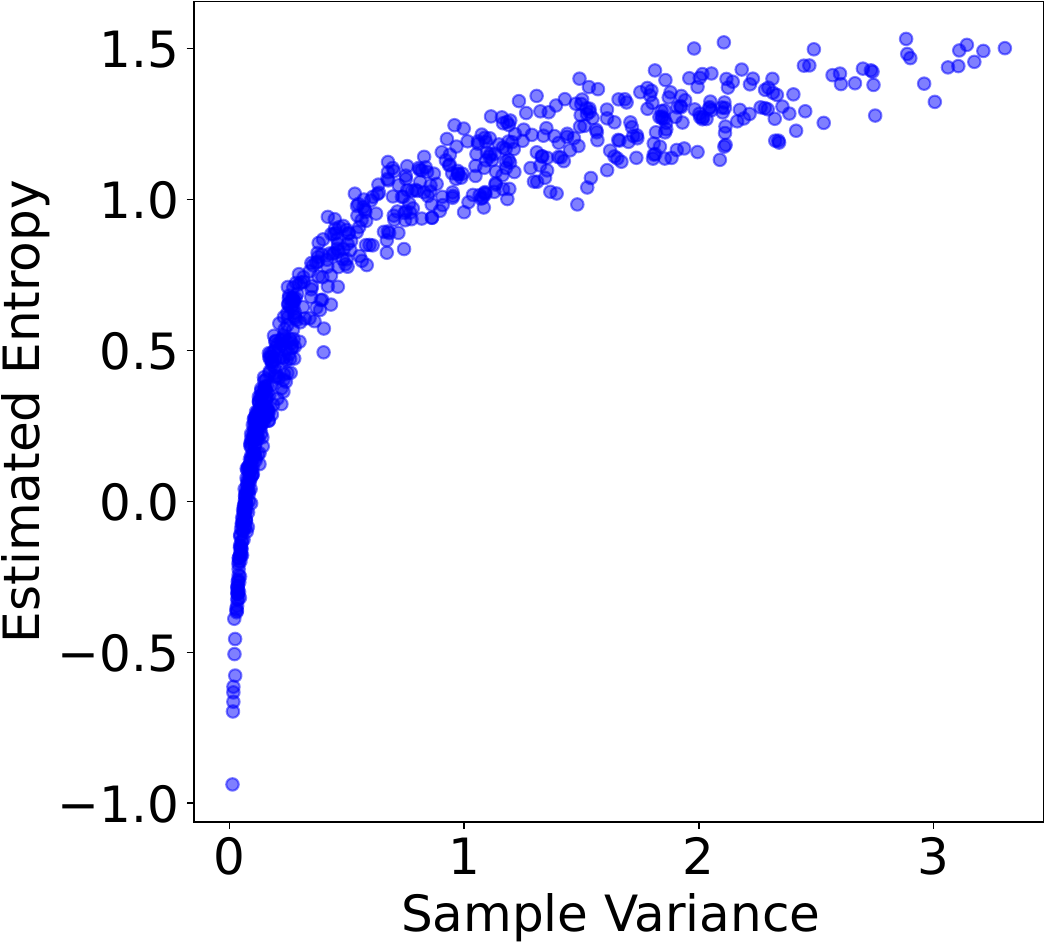}
        \caption{OT}
        \label{fig:fig3-right}
    \end{subfigure}
    \caption{The relationship between variance and information entropy for the 'HUFL' and 'OT' channels in ETTm1.}
    \label{fig: fig3}
\end{figure}

\subsection{Instance Normalization Formulations}
\label{sec: normalization formulations}
To address distribution shifts and mitigate the non-stationarity, instance normalization is commonly applied along the temporal dimension using a sliding window over time \cite{kim2021reversible, han2024sin, ye2024frequency}.
We illustrate this process using the simplified RevIN in the following \cite{liu2022non}.
Formally, given a sample \(\mathcal{S}_t=(\mathcal{X}_t, \mathcal{Y}_t)\), the normalization first computes the statistics of the history window for each channel, such as the mean \(\mu_{t,\text{x}}\) and standard deviation \(\sigma_{t,\text{x}}\).
Using these statistics, the input sequence \(\mathcal{X}_t\) is normalized by removing the mean and scaling by the variance as follows:
\begin{equation}
\label{eq: normalize}
\widetilde{\mathcal{X}}_t = \frac{\mathcal{X}_t - \mu_{t,\text{x}}}{\sigma_{t,\text{x}}}.
\end{equation}

\begin{figure*}[t]
\centering
\includegraphics[width=0.87\textwidth]{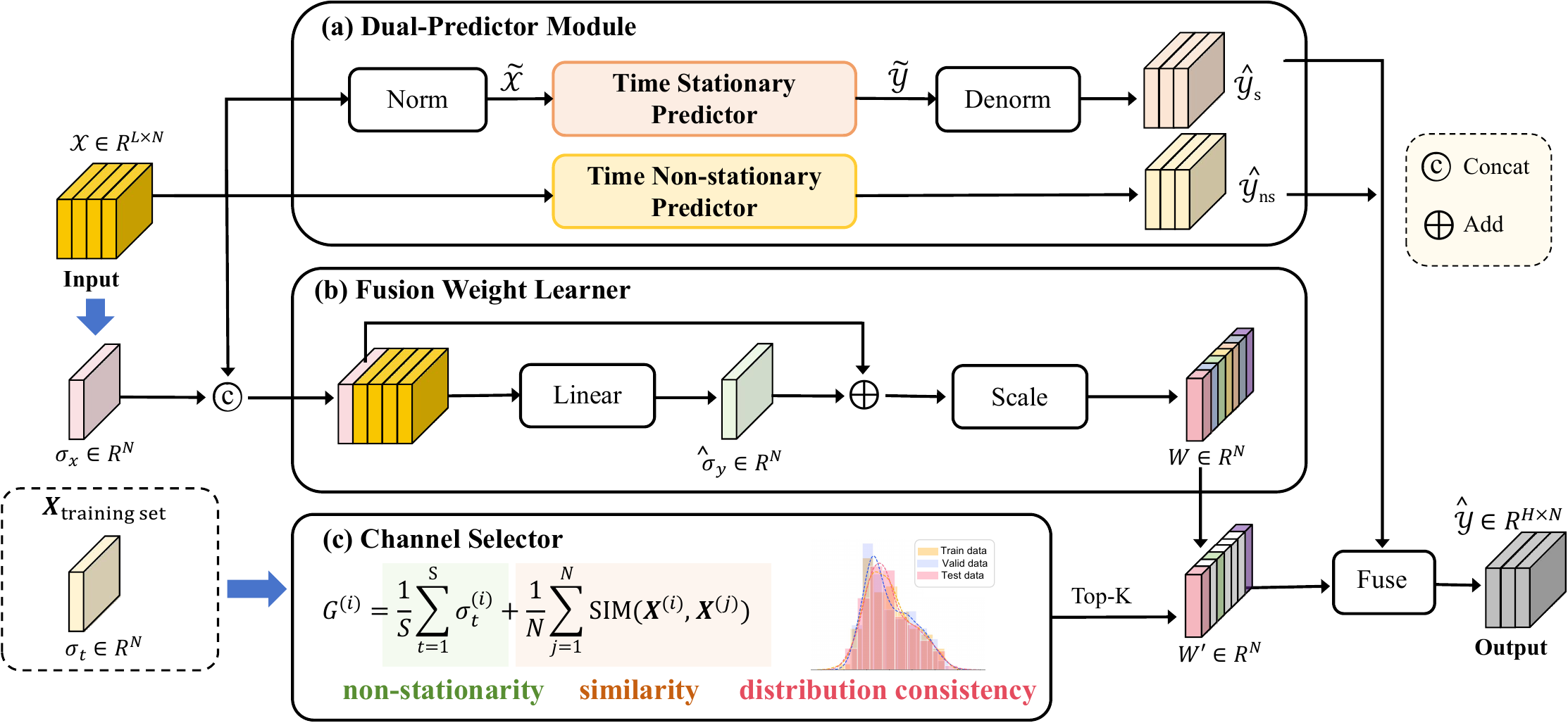} 
\vspace{2mm}
\caption{The architecture of CDFM consists of three key components: (1) The \textbf{Dual-Predictor Module} processes the input \(\mathcal{X}\) through two complementary branches to get the stationary prediction \(\hat{\mathcal{Y}_s}\) and the non-stationary prediction \(\hat{\mathcal{Y}_{ns}}\), respectively; (2) The \textbf{Fusion Weight Learner} dynamically learns fusion weight \(W\) of samples based on the historical standard deviation \(\sigma_{t,\mathbf{x}}\) and the predicted horizon standard deviation \(\hat{\sigma}_{t,\mathbf{y}}\); (3) The \textbf{Channel Selector} selects the top-k channels by evaluating their non-stationarity, similarity, and distributional consistency, and then recovers non-stationary information of these channels to obtain the final prediction \(\hat{\mathcal{Y}}\). }
\label{fig3}
\end{figure*}

Then, the forecasting model $\mathcal{F}$ predicts the normalized future values \(\widetilde{\mathcal{Y}}_t\) based on the normalized input \(\widetilde{\mathcal{X}}_t\):
\begin{equation}
\widetilde{\mathcal{Y}}_t = {\mathcal{F}}_{\Theta}(\widetilde{\mathcal{X}}_t).
\end{equation}

Lastly, we denormalize \(\widetilde{\mathcal{Y}}_t\) based on the statistics \(\mu_{t,\text{x}}\) and \(\sigma_{t,\text{x}}\) to obtain the final prediction \(\hat{\mathcal{Y}}_t\):
\begin{equation}
\label{eq: denormalize}
\hat{\mathcal{Y}}_t = \sigma_{t,\text{x}} \widetilde{\mathcal{Y}}_t  + \mu_{t,\text{x}}.
\end{equation}

\subsection{Quantifying Non-stationarity via Entropy}
\label{sec: Quantifying Non-stationarity via Information Entropy}

Shannon information theory provides a straightforward framework for studying non-stationary time processes, allowing for a quantitative characterization of non-stationarity in time series through information entropy \cite{granero2019information}.
This section consists of two key aspects: non-stationarity measure based on the Gaussian assumption and distribution-free non-stationarity estimation.
Our analysis demonstrates that variance can reflect information entropy of time series to some extent and it is a practical and interpretable measure for non-stationarity.

\textbf{(1) Non-stationarity measure via entropy based on the Gaussian assumption.}
Following previous works \cite{kim2021reversible, han2024sin}, we assume that the normalized input sequence follows a Gaussian distribution with mean \(0\) and standard deviation \(1\), i.e., \(\widetilde{\mathcal{X}}_t \sim N(0,1)\), as shown in Fig.~\ref{fig: fig2}~(a). Since the normalization is a linear transformation that preserves the properties of the Gaussian distribution, we can deduce that \(\mathcal{X}_t \sim N(\mu_{t,\text{x}},\sigma_{t,\text{x}}^2)\) as shown in Fig.~\ref{fig: fig2}~(b). 

For a Gaussian distribution \(N \sim (\mu,\sigma^2)\), the information entropy is given by \(H = \frac{1}{2}\ln (2\pi\sigma^2)\) \cite{tuan2007entropy}. Applying this formula, the entropy of \(\mathcal{X}_t\) is expressed as:
\begin{equation}
H(\mathcal{X}_t) = \frac{1}{2}\ln (2\pi \sigma_{t,\text{x}}^2).
\label{eq: H_X_t}
\end{equation}

Similarly, the information entropy of target sequence \(H(\mathcal{Y}_t)\) can be calculated based on the variance of the horizon window \(\sigma_{t,\text{y}}^2\).

Based on Gaussian assumption, the equation (\ref{eq: H_X_t}) reveals a direct proportionality between variance and information entropy. Actually, time series exhibit more complex and non-Gaussian distributions. In the following, empirical estimation of entropy based on Kernel Density Estimation (KDE) is introduced to measure non-stationarity without Gaussian assumption.

\textbf{(2) Non-stationarity estimation based on distribution-free KDE.}
To break through the limitations of Gaussian assumption, we conduct an empirical estimation of entropy \(H(\mathcal{X}_t)\) using KDE \cite{han2020optimal}. KDE avoids relying on predefined distributions, making it more flexible and broadly applicable to diverse and complex datasets \cite{han2020optimal}.
The relationship between variance and empirically estimated entropy is visualized in Fig.~\ref{fig: fig3}, and we observe a clear positive correlation between them. This confirms the effectiveness of variance as a practical and interpretable measure for non-stationarity.

\section{METHODOLOGY}

\subsection{Dual-Predictor Module}
The Dual-Predictor Module aims to improve time series forecasting by both enhancing the prediction of stationary components and extracting non-stationary representations from original series with complex distributions. 
This module comprises of two branches: the Time Stationary Predictor and the Time Non-stationary Predictor, which perform as complementary components to capture both stationary patterns and dynamic distribution changes.

\subsubsection{Time Stationary Predictor} 
To attenuate the non-stationarity of the original input sequence, we employ the simplified RevIN \cite{liu2022non} to align the input distribution, as illustrated in Section~\ref{sec: normalization formulations}. 
Following the equation (\ref{eq: normalize}), we obtain the normalized sequence \(\widetilde{\mathcal{X}}\).

We then adopt the linear model DLinear \cite{zeng2023transformers} as the Time Stationary Predictor to forecast the normalized future values:
\begin{equation}
\widetilde{\mathcal{Y}} = {\mathcal{F}}_{\Theta_s}(\widetilde{\mathcal{X}}),
\end{equation}
where \(\Theta_s\) denotes the parameters of the Time Stationary Predictor.

Finally, we denormalize the normalized predictions \(\widetilde{\mathcal{Y}}\) using the reverse translation and scaling operations defined in equation (\ref{eq: denormalize}) to obtain the final stationary prediction \(\hat{\mathcal{Y}_s}\).

In principle, the Time Stationary Predictor can adopt any advanced neural network architectures, offering flexibility and sufficient modeling capacity.

\subsubsection{Time Non-stationary Predictor}
To capture global temporal dependencies in the original series, we propose the Time Non-stationary Predictor, which uses the original sequence \(\mathcal{X}\) as input and produces the non-stationary prediction \(\hat{\mathcal{Y}}_{ns}\): 
\begin{equation}
\hat{\mathcal{Y}}_{ns} = {\mathcal{F}}_{\Theta_{ns}}(\mathcal{X}),
\end{equation}
where \(\Theta_{ns}\) represents the parameters of the Time Non-stationary Predictor.

Recent studies have shown that simple neural networks often outperform advanced ones on datasets with significant distribution shifts due to the avoidance of overfitting \cite{shao2024exploring, urettini2024gas}. 
Based on this observation, we utilize the linear model DLinear \cite{zeng2023transformers} as the Time Non-stationary Predictor.

\subsection{Fusion Weight Learner}
Due to the complexity of time series, samples often exhibit unique non-stationary characteristics. 
Guided by the analysis in Section~\ref{sec: Quantifying Non-stationarity via Information Entropy}, we adopt variance as a practical and interpretable measure to quantify the intrinsic non-stationarity.
Furthermore, greater entropy in the input sequence \(\mathcal{X}_t\) often indicates richer patterns and more complex non-stationary dynamics \cite{cao2004detecting}. 
Conversely, higher entropy in the output sequence \(\mathcal{Y}_t\) reflects increased uncertainty, which poses significant modeling challenges for forecasting \cite{yu2023exploring}.

To address these challenges and balance the trade-off between stationarity and non-stationarity, we propose a Fusion Weight Learner. This learner aims to dynamically recover the non-stationary information of different samples.
Specifically, this component generates fusion weight \(W\) for each sample by integrating the input variance \(\sigma_{\text{x}}\) and the predicted horizon variance \(\hat{\sigma}_{\text{y}}\), enabling the model to adaptively adjust weight based on the non-stationary dynamics and prediction difficulty.

Firstly, we employ an linear network with the sequence \(\mathcal{X}\) and its standard deviation \(\sigma_{\text{x}}\) as inputs, which enables the learner to capture both context and variance-related information to predict the standard deviation of the horizon window as follows:
\begin{equation}
\hat{\sigma}_{\text{y}} = \text{Linear}( \text{concat} (\sigma_{\text{x}}, \mathcal{X})).
\end{equation}

Then, the fusion weight \(W\) of each sample can be calculated based on \(\sigma_{\text{x}}\) and \(\sigma_{\text{y}}\):
\begin{equation}
    W = \lambda \odot (\sigma_{\text{x}} +  \hat{\sigma}_{\text{y}}),
\end{equation}
where \(\lambda \in \mathbb{R}^N\) is a learnable affine parameter to scale the standard deviation into suitable weight.

\subsection{Channel Selector}
The non-stationarity of different channels can significantly affect the forecasting performance of models. By selecting channels with higher non-stationary information, the model can better capture critical dynamic changes, thereby improving its adaptability and prediction accuracy. 
However, training on non-stationary sequences is challenging due to two main factors.
First, the non-stationarity of different channels can interact with each other. Selecting channels with conflicting or irrelevant non-stationary patterns may increase the data uncertainty and complexity, limiting the forecasting performance of models. 
Therefore, it is crucial to consider the similarity between channels, particularly for channel independent models. 
Second, when there is distribution shift between the training and testing sets, fusing non-stationary predictions can be risky, potentially leading to overfitting.
To address these challenges, the Channel Selector considers three key factors: non-stationarity, similarity, and distribution consistency of channels.

We measure the channel \(i\) based on its non-stationarity and similarity with other channels as follows:
\begin{equation}
\mathcal{G}(X^{(i)}) = \frac{1}{S}\sum\limits_{t=1}^{S}\sigma_t^{(i)} + \frac{\rho}{N}\sum\limits_{j=1}^{N}\text{SIM}(X^{(i)},X^{(j)}), 
\end{equation}
where \(S\) is the number of samples in the training set, \(\sigma_t^{(i)}\) is the standard deviation of sample \(\mathcal{S}_t^{(i)}\), \(\rho\) is a hyperparameter that balances the weight of non-stationarity and similarity, and \(\text{SIM}(X^{(i)},X^{(j)})\) represents the similarity between channels \(i\) and  \(j\), typically computed using a similarity measure like Pearson correlation. In the experiments, we set \(\rho=1\).

The indices of the top-k channels are formalized as:
\begin{equation}
\mathcal{C} = \underset{i}{\text{arg}} \> \text{Top-k}\{\mathcal{G}^{(i)} \mid i = 1, 2, \ldots, N\},
\end{equation}
where \(\mathcal{C}\) represents the selected channel indices, and we define \(\alpha\) as the selected channel ratio, \(\alpha = \lfloor {\frac{k}{N}}\rfloor\).

Considering the overfitting issues, we exclude channels that exhibit distributional shift between the training and testing sets by evaluating the change in loss on the validation set before and after fusion, with an example provided in 
 Appendix~\ref{sec: channel selector}. 

Finally, by selecting the specific channels based on non-stationarity, similarity, and distribution consistency, the model can focus on channels that exhibit significant non-stationary dynamics and minimize the impact of irrelevant or conflicting information.

The updated fusion weight \(W'\) are formalized as:

\begin{equation}
W' = \left\{
\begin{array}{ll}
W^{(i)}, & i \in \mathcal{C} \\
0, & i \notin \mathcal{C}  \\
\end{array},
\right.
\end{equation}
where \(\mathcal{C}\) represents the selected channel indices and  \(W^{(i)}\) represents the fusion weight for the channel \(i\).

Finally, \(W'\) is used to combine the stationary prediction \(\hat{\mathcal{Y}}_{s}\) and the non-stationary prediction \(\hat{\mathcal{Y}}_{ns}\) in a weighted sum:
\begin{equation}
\hat{\mathcal{Y}} = W' \odot \hat{\mathcal{Y}}_{ns}+(I-W') \odot \hat{\mathcal{Y}_s},
\end{equation}
where \(I\) is an all-one matrix, and \(\odot\) denotes the element-wise multiplication operation.

\begin{table}[t]
\caption{Summary of datasets.}
\centering
\large
\renewcommand{\arraystretch}{0.85}
\resizebox{0.45\textwidth}{!}{
\footnotesize   
\begin{tabular}{cccc}
\toprule
Datasets      &Sampling Frequency  & Variables     	     &
ADF Test Statistic	 \\ \midrule
ETTh1                             &1 Hour  &7        &-5.91   	\\ 
ETTh2                             &1 Hour &7         &-4.13    \\ 
ETTm1                             &15 Mins  &7            &-14.98  \\   
ETTm2		     	              &15 Mins   &7	         &-5.66    \\
Exchange	    	              &1 Day    &8		     &-1.90    \\  
Electricity	    	              &1 Hour	&321	     &-8.44    \\ 
Weather		               	      &10 Mins	&21 	     &-26.68   \\ \bottomrule
\end{tabular}}
\label{tab: tab1}
\end{table}

\begin{table*}[h]
\caption{Multivariate forecasting results with prediction lengths \(H \in \{96, 192, 336, 720\}\) and the input length \(L = 96\). Best results are highlighted in {\color[HTML]{FF0000} \textbf{red}} and the second {\textcolor{blue} {\ul underlined}}. Results on \textit{searched hyperparameter} input lengths are available in Appendix~\ref{sec: full results}.}
\centering
\large
\renewcommand{\arraystretch}{0.82}
\resizebox{0.97\textwidth}{!}{
\begin{tabular}{c|c|cc|cc|cc|cc|cc|cc|cc|cc|cc}
\toprule
\multicolumn{2}{c|}{Models}             & \multicolumn{2}{c|}{\begin{tabular}[c]{@{}c@{}}CDFM\\ (ours)\end{tabular}}     & \multicolumn{2}{c|}{\begin{tabular}[c]{@{}c@{}}iTransformer\\ (2024)\end{tabular}} & \multicolumn{2}{c|}{\begin{tabular}[c]{@{}c@{}}TimeMixer\\ (2024)\end{tabular}} & \multicolumn{2}{c|}{\begin{tabular}[c]{@{}c@{}}FITS\\ (2024)\end{tabular}} & \multicolumn{2}{c|}{\begin{tabular}[c]{@{}c@{}}PatchTST\\ (2023)\end{tabular}} & \multicolumn{2}{c|}{\begin{tabular}[c]{@{}c@{}}TimesNet\\ (2023a)\end{tabular}} & \multicolumn{2}{c|}{\begin{tabular}[c]{@{}c@{}}DLinear\\ (2023)\end{tabular}} & \multicolumn{2}{c|}{\begin{tabular}[c]{@{}c@{}}FEDformer\\ (2022)\end{tabular}} & \multicolumn{2}{c}{repeat}                                                 \\ \midrule
\multicolumn{2}{c|}{Metric}             & MSE                                   & MAE                                   & MSE                                     & MAE                                     & MSE                                    & MAE                                   & MSE                                 & MAE                                 & MSE                                   & MAE                                   & MSE                                                  & MAE                     & MSE                                                 & MAE                    & MSE                                    & MAE                                   & MSE                                & MAE                                   \\ \midrule
                                 & 96  & {\color[HTML]{FF0000} \textbf{0.374}} & {\color[HTML]{FF0000} \textbf{0.389}} & 0.387                                   & 0.405                                   & {\textcolor{blue} {\ul 0.378}}     & {\textcolor{blue} {\ul 0.397}}    & 0.395                               & 0.403                               & 0.413                                 & 0.419                                 & 0.389                                                & 0.412                   & 0.398                                               & 0.410                  & 0.385                                  & 0.425                                 & 1.294                              & 0.713                                 \\
                                 & 192 & {\color[HTML]{FF0000} \textbf{0.426}} & {\color[HTML]{FF0000} \textbf{0.419}} & 0.441                                   & 0.436                                   & 0.441                                  & 0.431    & 0.445                               & 0.432                               & 0.460                                 & 0.445                                 & 0.441                                                & 0.442                   & {\textcolor{blue} {\ul 0.434}}                  & {\textcolor{blue} {\ul 0.427}}                  & 0.441                                  & 0.461                                 & 1.325                              & 0.733                                 \\
                                 & 336 & {\color[HTML]{FF0000} \textbf{0.465}} & {\color[HTML]{FF0000} \textbf{0.438}} & 0.491                                   & 0.463                                   & 0.500                                  & {\textcolor{blue} {\ul 0.459}}    & {\textcolor{blue} {\ul 0.489}}  & 0.463                               & 0.497                                 & 0.463                                 & 0.491                                                & 0.467                   & 0.499                                               & 0.477                  & 0.491                                  & 0.473                                 & 1.330                              & 0.746                                 \\
                                 & 720 & {\color[HTML]{FF0000} \textbf{0.459}} & {\color[HTML]{FF0000} \textbf{0.455}} & 0.509                                   & 0.494                                   & {\textcolor{blue} {\ul 0.473}}     & {\textcolor{blue} {\ul 0.470}}    & 0.496                               & 0.485                               & 0.501                                 & 0.486                                 & 0.512                                                & 0.491                   & 0.508                                               & 0.503                  & 0.501                                  & 0.499                                 & 1.335                              & 0.755                                 \\ \cmidrule(rl){2-20}
\multirow{-5}{*}{\rotatebox{90}{ETTh1}}          & Avg & {\color[HTML]{FF0000} \textbf{0.431}} & {\color[HTML]{FF0000} \textbf{0.425}} & 0.457                                   & 0.450                                   & {\textcolor{blue} {\ul 0.448}}     & {\textcolor{blue} {\ul 0.439}}    & 0.456                               & 0.446                               & 0.468                                 & 0.453                                 & 0.458                                                & 0.453                   & 0.460                                               & 0.454                  & 0.455                                  & 0.465                                 & 1.321                              & 0.737                                 \\ \midrule
                                 & 96  & {\color[HTML]{FF0000} \textbf{0.285}} & {\color[HTML]{FF0000} \textbf{0.335}} & 0.301                                   & 0.350                                   & {\textcolor{blue} {\ul 0.295}}     & 0.346    & {\textcolor{blue} {\ul 0.295}}                              & {\textcolor{blue} {\ul 0.344}}  & 0.299                                 & 0.348                                 & 0.324                                                & 0.368                   & 0.315                                               & 0.374                  & 0.342                                  & 0.383                                 & 0.432                              & 0.422                                 \\
                                 & 192 & {\color[HTML]{FF0000} \textbf{0.366}} & {\color[HTML]{FF0000} \textbf{0.385}} & 0.380                                   & 0.399                                   & {\textcolor{blue} {\ul 0.376}}     & {\textcolor{blue} {\ul 0.395}}    & 0.382                               & 0.396                               & 0.383                                 & 0.398                                 & 0.393                                                & 0.410                   & 0.432                                               & 0.447                  & 0.434                                  & 0.440                                 & 0.534                              & 0.473                                 \\
                                 & 336 & {\color[HTML]{FF0000} \textbf{0.400}} & {\color[HTML]{FF0000} \textbf{0.414}} & 0.424                                   & 0.432                                   & 0.421                                  & 0.435                                 & {\textcolor{blue} {\ul 0.416}}  & {\textcolor{blue} {\ul 0.425}}  & 0.429                                 & 0.431                                 & 0.429                                                & 0.437                   & 0.486                                               & 0.481                  & 0.512                                  & 0.497                                 & 0.597                              & 0.511                                 \\
                                 & 720 & {\color[HTML]{FF0000} \textbf{0.389}} & {\color[HTML]{FF0000} \textbf{0.416}} & 0.430                                   & 0.447                                   & 0.445                                  & 0.458                                 & {\textcolor{blue} {\ul 0.418}}  & {\textcolor{blue} {\ul 0.437}}  & 0.424                                 & 0.445                                 & 0.433                                                & 0.448                   & 0.732                                               & 0.614                  & 0.467                                  & 0.476                                 & 0.594                              & 0.519                                 \\ \cmidrule(rl){2-20}
\multirow{-5}{*}{\rotatebox{90}{ETTh2}}          & Avg & {\color[HTML]{FF0000} \textbf{0.360}} & {\color[HTML]{FF0000} \textbf{0.388}} & 0.384                                   & 0.407                                   & 0.384                                  & 0.409                                 & {\textcolor{blue} {\ul 0.378}}  & {\textcolor{blue} {\ul 0.401}}  & 0.384                                 & 0.406                                 & 0.395                                                & 0.416                   & 0.491                                               & 0.479                  & 0.439                                  & 0.449                                 & 0.539                              & 0.481                                 \\ \midrule
                                 & 96  & {\color[HTML]{FF0000} \textbf{0.324}} & {\color[HTML]{FF0000} \textbf{0.357}} & 0.342                                   & 0.377                                   & {\textcolor{blue} {\ul 0.331}}     & {\textcolor{blue} {\ul 0.365}}    & 0.354                               & 0.375                               & {\textcolor{blue} {\ul 0.331}}                                 & 0.370                                 & 0.337                                                & 0.377                   & 0.346                                               & 0.374                  & 0.360                                  & 0.406                                 & 1.214                              & 0.665                                 \\
                                 & 192 & {\color[HTML]{FF0000} \textbf{0.367}} & {\color[HTML]{FF0000} \textbf{0.379}} & 0.383                                   & 0.396                                   & {\textcolor{blue} {\ul 0.370}}     & {\textcolor{blue} {\ul 0.385}}    & 0.392                               & 0.393                               & 0.374                                 & 0.395                                 & 0.395                                                & 0.406                   & 0.382                                               & 0.392                  & 0.395                                  & 0.427                                 & 1.261                              & 0.690                                 \\
                                 & 336 & {\color[HTML]{FF0000} \textbf{0.398}} & {\color[HTML]{FF0000} \textbf{0.400}} & 0.418                                   & 0.418                                   & {\textcolor{blue} {\ul 0.401}}     & {\textcolor{blue} {\ul 0.407}}    & 0.425                               & 0.415                               & 0.402                                 & 0.412                                 & 0.433                                                & 0.432                   & 0.414                                               & 0.414                  & 0.448                                  & 0.458                                 & 1.287                              & 0.707                                 \\
                                 & 720 & {\color[HTML]{FF0000} \textbf{0.456}} & {\color[HTML]{FF0000} \textbf{0.434}} & 0.487                                   & 0.457                                   & {\textcolor{blue} {\ul 0.462}}     & {\textcolor{blue} {\ul 0.442}}    & 0.486                               & 0.449                               & 0.466                                 & 0.446                                 & 0.484                                                & 0.458                   & 0.478                                               & 0.455                  & 0.491                                  & 0.479                                 & 1.322                              & 0.729                                 \\ \cmidrule(rl){2-20}
\multirow{-5}{*}{\rotatebox{90}{ETTm1}}          & Avg & {\color[HTML]{FF0000} \textbf{0.386}} & {\color[HTML]{FF0000} \textbf{0.393}} & 0.408                                   & 0.412                                   & {\textcolor{blue} {\ul 0.391}}     & {\textcolor{blue} {\ul 0.400}}    & 0.414                               & 0.408                               & 0.393                                 & 0.406                                 & 0.412                                                & 0.418                   & 0.405                                               & 0.409                  & 0.424                                  & 0.443                                 & 1.271                              & 0.698                                 \\ \midrule
                                 & 96  & 0.184                               & 0.268                                 & 0.186                                   & 0.272                                   & {\color[HTML]{FF0000} \textbf{0.175}}  & {\color[HTML]{FF0000} \textbf{0.257}} & 0.183                               & 0.266                               & {\textcolor{blue} {\ul 0.177}}    & {\textcolor{blue} {\ul 0.260}}    & 0.182                                                & 0.262                   & 0.184                                               & 0.276                  & 0.193                                  & 0.285                                 & 0.266                              & 0.328                                 \\
                                 & 192 & {\textcolor{blue} {\ul 0.244}}    & {\textcolor{blue} {\ul 0.303}}    & 0.254                                   & 0.314                                   & {\color[HTML]{FF0000} \textbf{0.237}}  & {\color[HTML]{FF0000} \textbf{0.300}} & 0.247                               & 0.305                               & 0.248                                 & 0.306                                 & 0.252                                                & 0.307                   & 0.282                                               & 0.357                  & 0.256                                  & 0.324                                 & 0.340                              & 0.371                                 \\
                                 & 336 & {\color[HTML]{FF0000} \textbf{0.302}} &  {\textcolor{blue} {\ul 0.341}}                                 & 0.316                                   & 0.351                                   & {\textcolor{blue} {\ul 0.303}}     & {\color[HTML]{FF0000} \textbf{0.340}} & 0.307                               & 0.342                               & {\textcolor{blue} {\ul 0.303}}                                 & {\textcolor{blue} {\ul 0.341}}    & 0.312                                                & 0.346                   & 0.324                                               & 0.364                  & 0.321                                  & 0.364                                 & 0.412                              & 0.410                                 \\
                                 & 720 & {\textcolor{blue} {\ul 0.404}}  & {\color[HTML]{FF0000} \textbf{0.396}} & 0.414                                   & 0.407                                   & {\color[HTML]{FF0000} \textbf{0.401}}      & {\textcolor{blue} {\ul 0.400}}    & 0.407                               & 0.401                               & 0.406                                 & 0.403                                 & 0.417                                                & 0.404                   & 0.441                                               & 0.454                  & 0.434                                  & 0.426                                 & 0.522                              & 0.466                                 \\ \cmidrule(rl){2-20}
\multirow{-5}{*}{\rotatebox{90}{ETTm2}}          & Avg & {\textcolor{blue} {\ul 0.284}}    & {\textcolor{blue} {\ul 0.327}}    & 0.293                                   & 0.336                                   & {\color[HTML]{FF0000} \textbf{0.279}}  & {\color[HTML]{FF0000} \textbf{0.324}} & 0.286                               & 0.329                               & {\textcolor{blue} {\ul 0.284}}                                 & {\textcolor{blue} {\ul 0.327}}                                 & 0.291                                                & 0.330                   & 0.308                                               & 0.363                  & 0.301                                  & 0.350                                 & 0.385                              & 0.394                                 \\ \midrule
                                 & 96  & 0.176                                 & 0.231                                 & 0.176                                   & 0.216                                   & {\color[HTML]{FF0000} \textbf{0.162}}  & {\color[HTML]{FF0000} \textbf{0.209}} & {\textcolor{blue} {\ul 0.167}}  & {\textcolor{blue} {\ul 0.214}}  & 0.177                                 & 0.219                                 & 0.168                                                & 0.218                   & 0.197                                               & 0.257                  & 0.236                                  & 0.325                                 & 0.259                              & 0.254                                 \\
                                 & 192 & {\textcolor{blue} {\ul 0.213}}    & 0.263                                 & 0.225                                   & {\textcolor{blue} {\ul 0.257}}      & {\color[HTML]{FF0000} \textbf{0.207}}  & {\color[HTML]{FF0000} \textbf{0.251}} & 0.215                               & {\textcolor{blue} {\ul 0.257}}                               & 0.225                                 & 0.259                                 & 0.226                                                & 0.267                   & 0.237                                               & 0.294                  & 0.268                                  & 0.337                                 & 0.309                              & 0.292                                 \\
                                 & 336 & {\color[HTML]{FF0000} \textbf{0.257}} & 0.300                                 & 0.281                                   & 0.299                                   & {\textcolor{blue} {\ul 0.264}}     & {\color[HTML]{FF0000} \textbf{0.294}} & 0.270                               & 0.299                               & 0.278                                 & {\textcolor{blue} {\ul 0.298}}    & 0.283                                                & 0.305                   & 0.283                                               & 0.332                  & 0.366                                  & 0.402                                 & 0.376                              & 0.338                                 \\
                                 & 720 & {\color[HTML]{FF0000} \textbf{0.317}} & {\color[HTML]{FF0000} \textbf{0.339}} & 0.358                                   & 0.350                                   & {\textcolor{blue} {\ul 0.345}}     & 0.348                                 & 0.347                               & {\textcolor{blue} {\ul 0.345}}  & 0.351                                 & 0.346                                 & 0.355                                                & 0.353                   & 0.347                                               & 0.382                  & 0.407                                  & 0.422                                 & 0.465                              & 0.394                                 \\ \cmidrule(rl){2-20}
\multirow{-5}{*}{\rotatebox{90}{Weather}}        & Avg & {\color[HTML]{FF0000} \textbf{0.241}} & 0.283                                 & 0.260                                   & 0.281                                   & {\textcolor{blue} {\ul 0.245}}     & {\color[HTML]{FF0000} \textbf{0.275}} & 0.250                               & {\textcolor{blue} {\ul 0.279}}  & 0.258                                 & 0.280                                 & 0.258                                                & 0.286                   & 0.266                                               & 0.316                  & 0.319                                  & 0.372                                 & 0.352                              & 0.320                                 \\ \midrule
                                 & 96  & {\color[HTML]{FF0000} \textbf{0.079}} & {\color[HTML]{FF0000} \textbf{0.196}} & 0.086                                   & 0.206                                   & 0.085                                  & 0.203                                 & 0.088                               & 0.210                               & 0.089                                 & 0.206                                 & 0.105                                                & 0.233                   & 0.089                                               & 0.219                  & 0.136                                  & 0.265                                 & {\textcolor{blue} {\ul 0.081}} & {\textcolor{blue} {\ul 0.196}}    \\
                                 & 192 & {\color[HTML]{FF0000} \textbf{0.156}} & {\textcolor{blue} {\ul 0.291}}    & 0.181                                   & 0.304                                   & 0.182                                  & 0.303                                 & 0.181                               & 0.304                               & 0.178                                 & 0.302                                 & 0.219                                                & 0.342                   & 0.180                                               & 0.319                  & 0.279                                  & 0.384                                 & {\textcolor{blue} {\ul 0.167}} & {\color[HTML]{FF0000} \textbf{0.289}} \\
                                 & 336 & {\color[HTML]{FF0000} \textbf{0.268}} & {\color[HTML]{FF0000} \textbf{0.389}} & 0.338                                   & 0.422                                   & 0.343                                  & 0.422                                 & 0.324                               & 0.413                               & 0.326                                 & 0.411                                 & 0.353                                                & 0.433                   & 0.313                                               & 0.423                  & 0.465                                  & 0.504                                 & {\textcolor{blue} {\ul 0.306}} & {\textcolor{blue} {\ul 0.398}}    \\
                                 & 720 & {\color[HTML]{FF0000} \textbf{0.653}} & {\color[HTML]{FF0000} \textbf{0.609}} & 0.853                                   & 0.696                                   & 1.059                                  & 0.767                                 & 0.846                               & 0.696                               & 0.840                                 & 0.690                                 & 0.912                                                & 0.724                   & 0.837                                               & 0.690                  & 1.169                                  & 0.826                                 & {\textcolor{blue} {\ul 0.810}} & {\textcolor{blue} {\ul 0.676}}    \\ \cmidrule(rl){2-20}
\multirow{-5}{*}{\rotatebox{90}{Exchange}} & Avg & {\color[HTML]{FF0000} \textbf{0.289}} & {\color[HTML]{FF0000} \textbf{0.371}} & 0.365                                   & 0.407                                   & 0.417                                  & 0.424                                 & 0.360                               & 0.406                               & 0.358                                 & 0.402                                 & 0.397                                                & 0.433                   & 0.355                                               & 0.413                  & 0.512                                  & 0.495                                 & {\textcolor{blue} {\ul 0.341}} & {\textcolor{blue} {\ul 0.390}}    \\ \midrule
                                 & 96  & 0.191                                 & 0.272                                 & {\color[HTML]{FF0000} \textbf{0.151}}   & {\color[HTML]{FF0000} \textbf{0.241}}   & {\textcolor{blue} {\ul 0.153}}     & {\textcolor{blue} {\ul 0.247}}    & 0.200                               & 0.278                               & 0.166                                 & 0.252                                 & 0.168                                                & 0.272                   & 0.195                                               & 0.277                  & 0.189                                  & 0.304                                 & 1.588                              & 0.945                                 \\
                                 & 192 & 0.191                                 & 0.278                                 & {\textcolor{blue} {\ul 0.167}}      & {\color[HTML]{FF0000} \textbf{0.258}}   & {\color[HTML]{FF0000} \textbf{0.166}}  & {\textcolor{blue} {\ul 0.258}}    & 0.200                               & 0.281                               & 0.174                                 & 0.261                                 & 0.186                                                & 0.289                   & 0.194                                               & 0.281                  & 0.198                                  & 0.312                                 & 1.596                              & 0.951                                 \\
                                 & 336 & 0.205                                 & 0.294                                 & {\color[HTML]{FF0000} \textbf{0.179}}   & {\color[HTML]{FF0000} \textbf{0.271}}   & {\textcolor{blue} {\ul 0.186}}     & {\textcolor{blue} {\ul 0.277}}    & 0.214                               & 0.295                               & 0.190                                 & 0.277                                 & 0.197                                                & 0.298                   & 0.207                                               & 0.296                  & 0.212                                  & 0.326                                 & 1.618                              & 0.961                                 \\
                                 & 720 & 0.242                                 & 0.327                                 & 0.229                                   & 0.319                                   & {\color[HTML]{FF0000} \textbf{0.225}}  & {\color[HTML]{FF0000} \textbf{0.312}} & 0.256                               & 0.328                               & 0.230                                 & {\textcolor{blue} {\ul 0.312}}    & {\textcolor{blue} {\ul 0.225}}                   & 0.322                   & 0.243                                               & 0.330                  & 0.242                                  & 0.351                                 & 1.647                              & 0.975                                 \\ \cmidrule(rl){2-20}
\multirow{-5}{*}{\rotatebox{90}{Electricity}}    & Avg & 0.207                                 & 0.293                                 & {\color[HTML]{FF0000} \textbf{0.182}}   & {\color[HTML]{FF0000} \textbf{0.272}}   & {\textcolor{blue} {\ul 0.183}}     & {\textcolor{blue} {\ul 0.274}}    & 0.218                               & 0.296                               & 0.190                                 & 0.276                                 & 0.194                                                & 0.295                   & 0.210                                               & 0.296                  & 0.210                                  & 0.323                                 & 1.612                              & 0.958                                 \\\midrule
\rowcolor{pink!50}
\multicolumn{2}{c}{1st Count}          & {\color[HTML]{FF0000} \textbf{24}}    & {\color[HTML]{FF0000} \textbf{21}}    & 3                                       & 4                                       & {\textcolor{blue} {\ul 8}}         & {\textcolor{blue} {\ul 9}}        & 0                                   & 0                                   & 0                                     & 0                                     & 0                                                    & 0                       & 0                                                   & 0                      & 0                                      & 0                                     & 0                                  & 1                      \\ \bottomrule  
\end{tabular}}
\label{tab: tab2_all}
\end{table*}

\section{Experiments}

\subsection{Experimental Setup}

\subsubsection{Datasets}
We conduct multivariate time series forecasting experiments on seven widely-used time series datasets, including ETTh1, ETTh2, ETTm1, ETTm2, Weather, Exchange Rate and Electricity.
In addition, we adopt the ADF \cite{elliott1992efficient} test statistic to quantify the degree of non-stationarity of all dataset. 
A larger ADF test statistic indicates a higher degree of non-stationarity. 
Table~\ref{tab: tab1} summarizes the primary statistics of the datasets. 
We can see that Exchange Rate, ETTh2, and ETTm2 exhibit relatively high non-stationarity, indicating challenges for forecasting. 

\subsubsection{Baselines}
We establish a comprehensive set of baselines to evaluate our approach.
\begin{itemize}[left=0em]
\item For overall performance comparison, we carefully choose seven state-of-the-art (SOTA) forecasting models: MLP-based models: DLinear \cite{zeng2023transformers}, FITS \cite{xu2024fits} and TimeMixer \cite{wang2024timemixer}. Transformer-based models: iTransformer \cite{liu2024itransformer}, PatchTST \cite{nie2023time} and FEDformer \cite{zhou2022fedformer}. CNN-based models: TimesNet \cite{wu2023timesnet}.  Besides, we include a naive method: Repeat, which repeats the last value in the historical window.
\item For comparison with non-stationary models, we consider four SOTA methods: TFPS \cite{sun2024learning}, Koopa \cite{liu2023koopa}, NS-Trans\cite{liu2022non} and HTV-Trans \cite{wang2024considering}. 
\item For comparison with normalization methods, we include five existing normalization techniques: RevIN \cite{kim2021reversible}, Dish-TS \cite{fan2023dish}, SAN \cite{liu2024adaptive}, SIN \cite{han2024sin} and FAN \cite{ye2024frequency}.
\end{itemize}

\subsubsection{Experimental Setup}

CDFM was implemented by Pytorch \cite{paszke2019pytorch} and trained on a single NVIDIA RTX 3090 24GB GPU. More experimental details are provided in Appendix~\ref{sec: experimental setup}.

\begin{table*}[t]
\caption{Comparison between CDFM and non-stationary models with prediction lengths \(H \in
\{96, 192, 336, 720\}\) and the input length \(L=96\). 
Full results are available in Appendix~\ref{sec: full results}.
}
\centering
\large
\renewcommand{\arraystretch}{0.80}
\resizebox{0.98\textwidth}{!}{
\begin{tabular}{c|cc|cc|cc|cc|cc|cc|cc}
\toprule
Dataset         & \multicolumn{2}{c|}{ETTh1}                                                     & \multicolumn{2}{c|}{ETTh2}                                                     & \multicolumn{2}{c|}{ETTm1}                                                     & \multicolumn{2}{c}{ETTm2}                                                     & \multicolumn{2}{c|}{Weather}                                                   & \multicolumn{2}{c|}{Exchange\_rate}                                            & \multicolumn{2}{c}{Electricity}                                               \\ \midrule
Metric          & MSE                                   & MAE                                   & MSE                                   & MAE                                   & MSE                                   & MAE                                   & MSE                                   & MAE                                   & MSE                                   & MAE                                   & MSE                                   & MAE                                   & MSE                                   & MAE                                   \\ \midrule
NS-Trans(2022)  & 0.679                                 & 0.563                                 & 0.502                                 & 0.475                                 & 0.524                                 & 0.468                                 & 0.617                                 & 0.484                                 & 0.282                                 & 0.307                                 & 0.546                                 & 0.490                                 & 0.193                                 & 0.296                                 \\
Koopa(2023)     & 0.454                                 & 0.443                                 & 0.387                                 & 0.412                                 & 0.399                                 & {\textcolor{blue} {\ul  0.406}}    & {\textcolor{blue} {\ul  0.284}}    & 0.329                                 & {\textcolor{blue} {\ul  0.245}}    & {\textcolor{blue} {\ul  0.272}}    & {\textcolor{blue} {\ul  0.382}}    & 0.419                                 & {\textcolor{blue} {\ul  0.189}}    & {\textcolor{blue} {\ul  0.283}}    \\
HTV-Trans(2024) & 0.457                                 & {\textcolor{blue} {\ul  0.440}}    & 0.431                                 & 0.431                                 & 0.406                                 & 0.406                                 & 0.288                                 & 0.329                                 & 0.258                                 & 0.284                                 & 0.399                                 & 0.429                                 & 0.214                                 & 0.310                                 \\
TFPS(2024)      & {\textcolor{blue} {\ul  0.448}}    & 0.443                                 & {\textcolor{blue} {\ul  0.380}}    & {\textcolor{blue} {\ul  0.403}}    & {\textcolor{blue} {\ul  0.395}}    & 0.407                                 & {\color[HTML]{FF0000} \textbf{0.276}} & {\color[HTML]{FF0000} \textbf{0.321}} & {\color[HTML]{FF0000} \textbf{0.241}} & {\color[HTML]{FF0000} \textbf{0.271}} & 0.395                                 & {\textcolor{blue} {\ul  0.414}}    & {\color[HTML]{FF0000} \textbf{0.183}} & {\color[HTML]{FF0000} \textbf{0.280}} \\ \midrule
CDFM(ours)      & {\color[HTML]{FF0000} \textbf{0.431}} & {\color[HTML]{FF0000} \textbf{0.425}} & {\color[HTML]{FF0000} \textbf{0.360}} & {\color[HTML]{FF0000} \textbf{0.388}} & {\color[HTML]{FF0000} \textbf{0.386}} & {\color[HTML]{FF0000} \textbf{0.393}} & {\textcolor{blue} {\ul  0.284}}    & {\textcolor{blue} {\ul  0.327}}    & {\color[HTML]{FF0000} \textbf{0.241}} & 0.283                                 & {\color[HTML]{FF0000} \textbf{0.289}} & {\color[HTML]{FF0000} \textbf{0.371}} & 0.207                                 & 0.293 \\ \bottomrule
\end{tabular}}

\label{tab: tab4}
\end{table*}

\begin{table*}[t]
\caption{Performance promotion by applying CDFM to other models with \(L=96\), with full results available in Appendix~\ref{sec: full results}.}
\centering
\large
\renewcommand{\arraystretch}{0.80}
\resizebox{0.92\textwidth}{!}{

\begin{tabular}{c|cccc|c|cccc|c|cccc|c}
\toprule
Dataset      & \multicolumn{5}{c|}{ETTh2}                                                                                                                                                                             & \multicolumn{5}{c|}{Weather}                                                                                                                                                                           & \multicolumn{5}{c}{Electricity}                                                                                                                                                                       \\ \midrule
Horizon      & 96                                    & 192                                   & 336                                   & 720                                   & Avg               & 96                                    & 192                                   & 336                                   & 720                                   & Avg                                   & 96                                    & 192                                   & 336                                   & 720                                   & Avg           \\ \midrule
TimeMixer    & 0.295                                 & 0.376                                 & 0.421                                 & 0.445                                 & 0.384                                 & 0.162                                 & 0.207                                 & 0.264                                 & 0.345                                 & 0.245                                 & 0.153                                 & 0.166                                 & 0.186                                 & 0.225                                 & 0.183                                 \\
+CDFM         & {\color[HTML]{FF0000} \textbf{0.286}} & {\color[HTML]{FF0000} \textbf{0.368}} & {\color[HTML]{FF0000} \textbf{0.412}} & {\color[HTML]{FF0000} \textbf{0.428}} & {\color[HTML]{FF0000} \textbf{0.374}} & {\color[HTML]{FF0000} \textbf{0.160}} & {\color[HTML]{FF0000} \textbf{0.201}} & {\color[HTML]{FF0000} \textbf{0.251}} & {\color[HTML]{FF0000} \textbf{0.323}} & {\color[HTML]{FF0000} \textbf{0.234}} & {\color[HTML]{FF0000} \textbf{0.152}} & {\color[HTML]{FF0000} \textbf{0.164}} & {\color[HTML]{FF0000} \textbf{0.181}} & {\color[HTML]{FF0000} \textbf{0.218}} & {\color[HTML]{FF0000} \textbf{0.179}} \\ \midrule
iTransformer & 0.301                                 & 0.380                                 & 0.424                                 & 0.430                                 & 0.384                                 & 0.176                                 & 0.225                                 & 0.281                                 & 0.358                                 & 0.260                                 & 0.151                                 & 0.167                                 & 0.179                                 & 0.229                                 & 0.182                                 \\
+CDFM         & {\color[HTML]{FF0000} \textbf{0.299}} & {\color[HTML]{FF0000} \textbf{0.379}} & {\color[HTML]{FF0000} \textbf{0.420}} & {\color[HTML]{FF0000} \textbf{0.409}} & {\color[HTML]{FF0000} \textbf{0.377}} & {\color[HTML]{FF0000} \textbf{0.165}} & {\color[HTML]{FF0000} \textbf{0.213}} & {\color[HTML]{FF0000} \textbf{0.263}} & {\color[HTML]{FF0000} \textbf{0.331}} & {\color[HTML]{FF0000} \textbf{0.243}} & {\color[HTML]{FF0000} \textbf{0.143}} & {\color[HTML]{FF0000} \textbf{0.159}} & {\color[HTML]{FF0000} \textbf{0.172}} & {\color[HTML]{FF0000} \textbf{0.198}} & {\color[HTML]{FF0000} \textbf{0.168}}
\\ \bottomrule
\end{tabular}}
\label{tab: tab5}
\vspace{2mm}
\end{table*}
\begin{table}[!t]
\caption{Comparison between CDFM and normalization approaches with \(L=96\), full results in Appendix~\ref{sec: full results}. 
}
\centering
\large

\renewcommand{\arraystretch}{0.78}
\resizebox{0.37\textwidth}{!}{
\begin{tabular}{c|cc|cc}
\toprule
                         & \multicolumn{2}{c|}{ETTh1}                                                     & \multicolumn{2}{c}{ETTm1}                                                     \\ \cmidrule(rl){2-5}
\multirow{-2}{*}{Models} & MSE                                   & MAE                                   & MSE                                   & MAE                                   \\ \midrule
DLinear                  & 0.460                                 & 0.454                                 & 0.405                                 & 0.409                                 \\
+RevIN                    & {\textcolor{blue} {\ul 0.442}}    & {\textcolor{blue} {\ul 0.429}}    & 0.410                                 & {\textcolor{blue} {\ul 0.400}}    \\

+Dish-TS                  & 0.461                                 & 0.449                                 & 0.406                                 & 0.410                                 \\
+FAN                      & 0.477                                 & 0.463                                 & 0.401                                 & 0.414                                 \\
+SAN                      & 0.459                                 & 0.452                                 & {\textcolor{blue} {\ul 0.394}}    & 0.410                                 \\
+SIN                      & 0.450                                 & 0.444                                 & 0.405                                 & 0.412    
\\ \midrule
+CDFM                     & {\color[HTML]{FF0000} \textbf{0.431}} & {\color[HTML]{FF0000} \textbf{0.425}} & {\color[HTML]{FF0000} \textbf{0.386}} & {\color[HTML]{FF0000} \textbf{0.393}} \\

\bottomrule
\end{tabular}}
\label{tab: tab3}
\end{table}

\begin{table*}[t]
\caption{Ablation study of CDFM components. ‘\(\alpha=1\)’ refers to including all channels in the fusion process. 'Static Weight' refers a trainable static matrix shared across all samples.}
\centering
\footnotesize
\large

\renewcommand{\arraystretch}{0.80}
\resizebox{0.96\textwidth}{!}{
\begin{tabular}{c|c|c|c|cccc|cccc}
\toprule
\multicolumn{2}{c|}{Dual-predictor}                                            &                                                                                      &                          & \multicolumn{4}{c|}{ETTh1}                                                                                                                                     & \multicolumn{4}{c}{ETTh2}                                                                                                                                     \\ \cmidrule(rl){1-2} \cmidrule(rl){5-12}
Stationary Predictor &Non-stationary Predictor & \multirow{-2}{*}{\begin{tabular}[c]{@{}c@{}}Channel \\ Selector\end{tabular}} & \multirow{-2}{*}{\begin{tabular}[c]{@{}c@{}}Dynamic Weight \\ Learner\end{tabular}}  & 96                                    & 192                                   & 336                                   & 720                                   & 96                                    & 192                                   & 336                                   & 720                                   \\ \midrule
\checkmark                         &                               &                                    &                                           & 0.384                                 & 0.436                                 & 0.479                                 & 0.481                                 & 0.289                                 & 0.375                                 & 0.417                                 & 0.420                                  \\ \midrule
                          & \checkmark                             &                                    &                                           & 0.398                                 & 0.434                                 & 0.499                                 & 0.508                                 & 0.315                                 & 0.432                                 & 0.486                                 & 0.732                                 \\ \midrule
\checkmark                         & \checkmark                             & \checkmark                                  & Static Weight                             & 0.379                                 & 0.428                                 & 0.467                                 & 0.461                                 & 0.286                                 & 0.369                                 & 0.402                                 & 0.392                                 \\ \midrule
\checkmark                         & \checkmark                             & \(\alpha=1 \)                           & \checkmark                                         & 0.380                                 & 0.430                                 & 0.478                                 & 0.488                                 & 0.287                                 & 0.376                                 & 0.405                                 & 0.458                                 \\ \midrule
\checkmark                         & \checkmark                             & \checkmark                                  & \checkmark                                         & {\color[HTML]{FF0000} \textbf{0.374}} & {\color[HTML]{FF0000} \textbf{0.426}} & {\color[HTML]{FF0000} \textbf{0.465}} & {\color[HTML]{FF0000} \textbf{0.459}} & {\color[HTML]{FF0000} \textbf{0.285}} & {\color[HTML]{FF0000} \textbf{0.366}} & {\color[HTML]{FF0000} \textbf{0.400}} & {\color[HTML]{FF0000} \textbf{0.389}} \\ \bottomrule
\end{tabular}}

\label{tab: tab6}
\end{table*}

\subsection{Overall Performance Comparison}

Table~\ref{tab: tab2_all} presents comprehensive forecasting results, which demonstrate that CDFM achieves superior forecasting performance.
Notably, CDFM demonstrates remarkable performance on datasets with high non-stationarity. For instance, it outperforms all other models on the highly non-stationary ETTh2 datasets, achieving MSE reductions of 11.2\%.
While competing methods like TimeMixer and FITS show competitive results across several datasets, CDFM still outperforms them, showing an average reduction of 4.5\% in MSE and 2.2\% in MAE.
These significant improvements can be attributed to CDFM's ability to effectively integrate both stationary and non-stationary information.
Furthermore, its innovative dynamic fusion mechanism enables dynamic recovery of critical non-stationary information.
By restoring non-stationary information to capture global temporal dependencies, CDFM achieves more precise predictions compared to other baseline models.

For the Electricity dataset, CDFM is inferior to iTransformer and TimeMixer, primarily due to the inherent limitations of the fixed short input length, which restricts the number of parameters of CDFM. When we search for the best input length \(L=\{96, 336, 512\}\), CDFM exhibits SOTA performance.
Full results are available in Appendix~\ref{sec: full results}.

\subsection{Comparison with Non-stationary Models }
Non-stationarity methods such as TFPS \cite{sun2024learning}, Koopa \cite{liu2023koopa}, NS-Trans \cite{liu2022non}, and HTV-Trans \cite{wang2024considering} can utilize the non-stationarity information to enhance performance 
and are widely used for non-stationary time series forecasting. 
We compare our CDFM with these SOTA non-stationarity methods and Table~\ref{tab: tab4} presents the performance comparison. 
We see that CDFM outperforms other non-stationary models in most cases.
By accounting for the distinct non-stationarity of different samples and dynamically integrating them with stationary features, CDFM effectively enhances prediction accuracy.

Notably, based on lightweight design, CDFM achieves competitive performance while maintaining a lower computational cost compared to more complex transformer-based models like TFPS and HTV-Trans shown in Fig.~\ref{fig: fig9}.

\subsection{Comparison with Normalization Methods}
We compare CDFM's performance with existing normalization methods, reporting the evaluation on DLinear in Table~\ref{tab: tab3}.
The results demonstrate that CDFM achieves SOTA performance, with a significant average MSE reduction of 4.8\%. We attribute this improvement to CDFM's ability to incorporate non-stationary information, contrasting with models that depend solely on normalization to attenuate the non-stationarity of original series.
Although normalization methods provide stable data distribution for models, excessive reliance on them can result in over-smoothed predictions, limiting the ability of models to effectively handle real-world time series with complex and diverse distributions. 

\subsection{Framework Generality}
Actually, as for our proposed framework CDFM, the Time Stationary Predictor can be replaced by any other backbone.
To demonstrate the generality of our proposed framework, we integrate it with TimeMixer and iTransformer and report the performance improvements in Table~\ref{tab: tab5}.
The results reveal that our framework consistently enhances the performance of TimeMixer and iTransformer, achieving an average reduction of 2.8\% and 1.7\% on ETTh2, 3.9\% and 6.4\% on Weather and 1.9\% and 6.9\% on Electricity, respectively.
These findings validate that CDFM is not only effective but also highly adaptable, serving as a lightweight and flexible framework that enhances the non-stationary prediction performance of other models with negligible impact on their computational complexity.

\begin{figure}[t]
\centering
\includegraphics[width=0.45\textwidth]{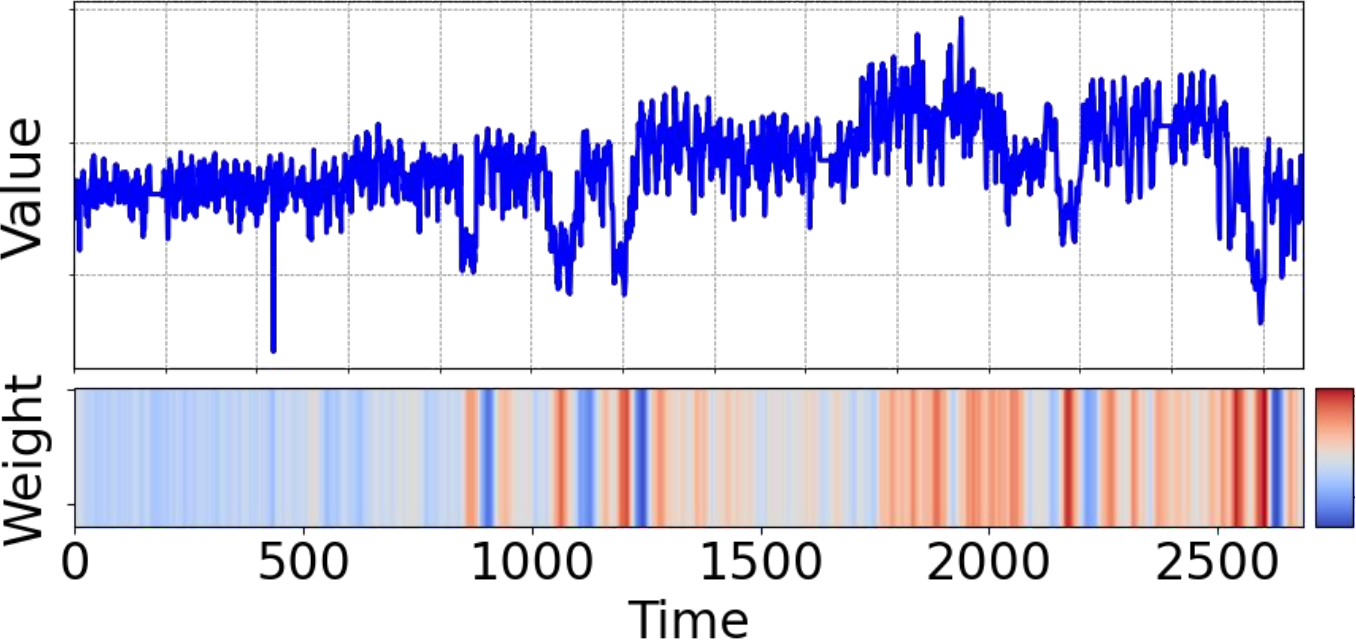} 
\caption{The dynamic fusion weight of input-96-predict-192 on the ETTh2 dataset. The highly non-stationary samples have large weights.}
\label{fig: fig6}
\vspace{-2mm}
\end{figure}

\subsection{Ablation Study} 
To assess the significance of each module of the CDFM, we conduct ablation studies using various model variants. These variants combine different configurations of Dual-Predictor, Channel Selector, and Fusion Weight Learner. The results are presented in Table~\ref{tab: tab6}.
\begin{itemize}[left=0em]
\item \textbf{Best Result.} The full CDFM model, i.e., including Dual-Predictor, Channel Selector, and Fusion Weight Learner, achieves the best performance across all forecast horizons. 
\item \textbf{Time Stationary Predictor vs. Time Non-stationary Predictor.} 
We find that both Stationary Predictor and Non-stationary Predictor alone yield worse performance, and the Stationary Predictor performs better due to the stable input distribution.
\item \textbf{Impact of Fusion Weight Learner.}
Replacing the dynamic fusion weight with static fusion weight leads to a decline in performance, which underscores the importance of dynamically recovering intrinsic non-stationary information for each sample.
\item \textbf{Importance of Channel Selector.}
When comparing selective channel fusion (\(\alpha<1\)) with full-channel fusion (\(\alpha=1\)), the selective strategy improves the robustness of the model trained on non-stationary sequences and shows better prediction performance.

\end{itemize}
These results validate the necessity of each module in the CDFM framework, demonstrating how their combined contributions leads to superior forecasting performance.

\section{Analysis}
In this section, we analyze the fusion weight and evaluate the model's efficiency.
Furthermore, we investigate the effect of the selected channel ratio \(\alpha\) and the input length \(L\) in Appendix~\ref{sec: parameter sensitivity}.

\subsection{Fusion Weight Visualization} 
To validate that CDFM can dynamically adjust the fusion weight based on the non-stationarity of the samples, we present the visualization of the fusion weight on the ETTh2 dataset in Fig.~\ref{fig: fig6}. 
It is evident that the fusion weight exhibits continuous variations over time, closely reflecting the fluctuations that indicate dynamic changes in the data.
This observation suggests that CDFM effectively captures the temporal dynamics of the data and adjusts the fusion weight accordingly. The dynamic adjustment of fusion weight enhances the adaptability of the model to changes in data distribution, thereby improving its robustness and prediction accuracy.

\begin{figure}[t]
    \centering
    \begin{subfigure}{0.22\textwidth}
        \centering
        \includegraphics[width=\linewidth]{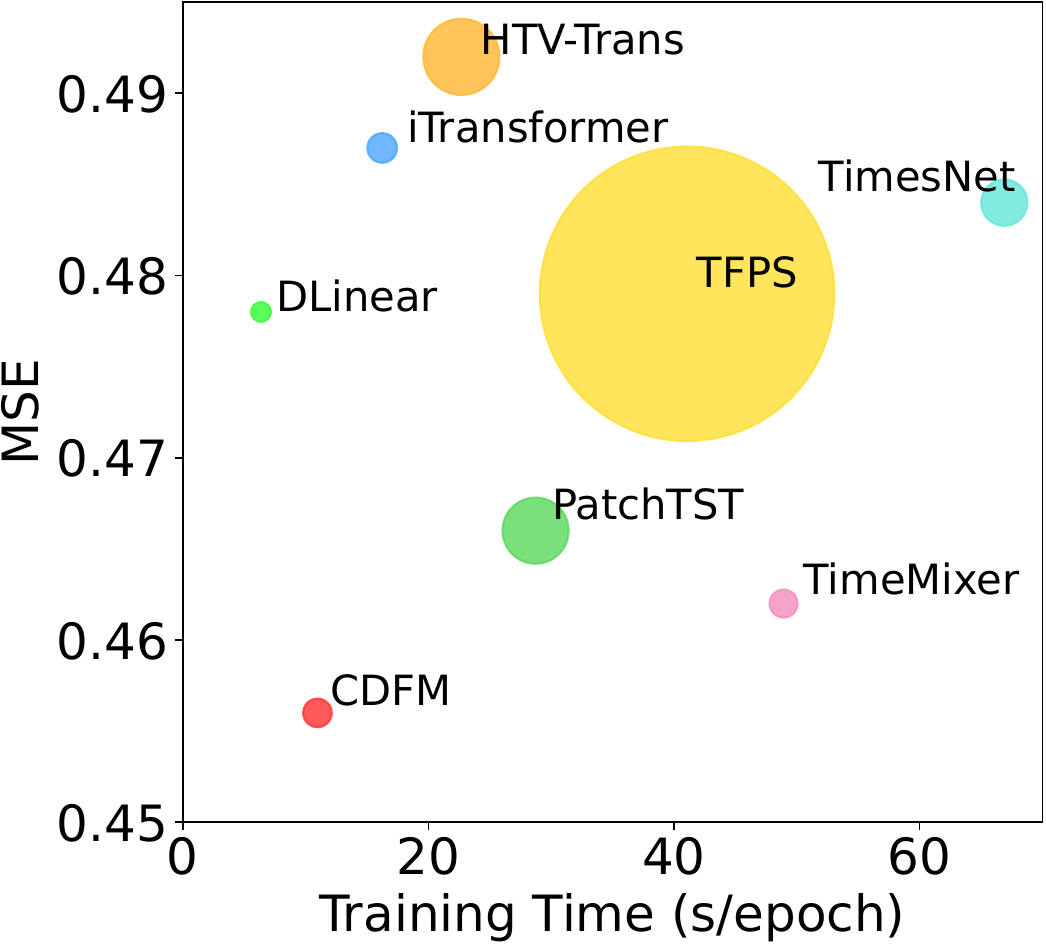}
        \caption{ETTm1}
        \label{fig:fig9-left}
    \end{subfigure}
    \hspace{0.02\textwidth}
    \begin{subfigure}{0.22\textwidth}
        \centering
        \includegraphics[width=\linewidth]{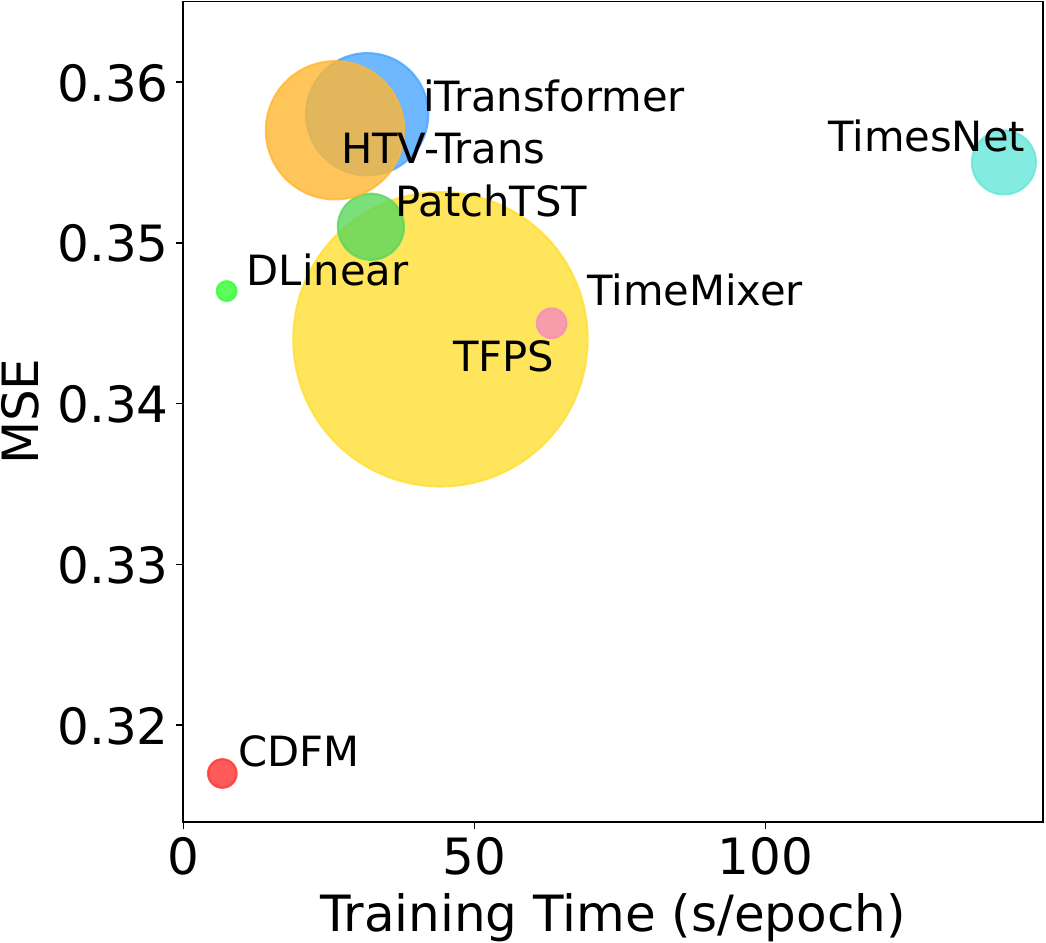}
        \caption{Weather}
        \label{fig:fig9-right}
    \end{subfigure}
    \caption{Model efficiency comparison across different models of input-96-predict-720 on ETTm1 and Weather.}
    \label{fig: fig9}
\vspace{-2mm}
\end{figure}

\subsection{Model Efficiency} 
We evaluate the efficiency of CDFM in terms of prediction performance, training speed and memory usage, as shown in Fig.~\ref{fig: fig9}.
Take the Weather dataset as an example: (1) CDFM achieves an MSE of 0.317, outperforming the other baselines; (2) CDFM runs in 6.66 s, making it 4.8× faster than PatchTST (32.19 s) and 9.5× faster than TimeMixer (63.25 s).
(3) CDFM uses 0.27 MB of GPU memory, significantly less than iTransformer (4.92 MB) and comparable to TimeMixer (0.29 MB). 
Therefore, CDFM achieves the best balance between performance and efficiency.

\section{Conclusion}

This paper addresses the issues posed by non-stationarity in time series. Unlike previous studies that simply attenuate non-stationarity by instance normalization, we propose a novel and lightweight channel-wise dynamic fusion forecasting model.
Our CDFM framework consists of Dual-Predictor Module, Channel Selector and Fusion Weight Learner.
CDFM selectively and dynamically leverages intrinsic non-stationarity in original series for forecasting while still keeping the predictability of normalized series.
Comprehensive experiments on seven datasets demonstrate CDFM's outstanding forecasting performance.
Furthermore, CDFM also achieves competitive performance compared with both existing non-stationary models and normalization methods, indicating its superiority in utilizing the non-stationarity to enhance the forecasting performance.
In future work, we will investigate the effect of instance normalization on variable correlation, which is crucial for channel-dependent models.

\bibliographystyle{ACM-Reference-Format}
\bibliography{main}

\appendix
\section{Deep Models for Time Series Forecasting}
\label{sec: deep models for time series forecasting}
Deep learning models for time series forecasting have been extensively studied in recent decades due to the powerful representation capability of neural networks. 

\textbf{RNN-based models} are designed for sequence modeling based on the Markov assumption \cite{yunpeng2017multi, Rangapuram2018deep, hu2020time}. But their recursive inference framework inherently struggles with capturing long-term dependencies due to issues like vanishing gradients, which makes them less effective for modeling complex and long-term time series.

On the other hand, \textbf{CNN-based models}, although effective in capturing local patterns, suffer from modeling long-term dependency due to limited convolutional reception field. Temporal Convolutional Networks (TCN) address this limitation by expanding the receptive field \cite{bai2018empirical}. Recent works have applied TCN methods to long-term series forecasting tasks including SCINet \cite{liu2022scinet}, TimesNet \cite{wu2023timesnet} and ModernTCN \cite{luo2024moderntcn}. 

Recently, \textbf{Transformer-based models} have achieved superior performance in time series forecasting due to their ability to capture long-term dependencies via the attention mechanism \cite{zhou2021informer, wu2021autoformer, zhang2023crossformer}.
More recent advancements, such as Fedformer \cite{zhou2022fedformer}, Non-stationary Transformers \cite{liu2022non}, PatchTST \cite{nie2023time} and iTransformer \cite{liu2024itransformer}, have modified Vanilla Transformer to better address the unique challenges of multivariate time series forecasting, such as handling complex temporal dependencies, non-stationarity, long-term patterns and multivariate correlations.

In contrast to the complex architectures of Transformers, DLinear \cite{zeng2023transformers} has highlighted that simple \textbf{MLP-based models} can achieve competitive performance compared to Transformer-based models.
Subsequently, many lightweight models have been explored and proposed, including RLinear \cite{li2023revisiting}, FITS\cite{xu2024fits}, SparseTSF \cite{lin2024sparsetsf} and TimeMixer \cite{wang2024timemixer}. 

Furthermore, the state space model Mamba has received attention since its ability of sequence modeling while maintaining near-linear complexity \cite{wang2025mamba, ahamed2024timemachine}.

Recent studies have shown that pretrained Large Language Models (\textbf{LLM-based Models}) can also achieve remarkable performance in time series forecasting tasks. 
Fine-tuning these pretrained models for time series analysis has yielded promising results, such as LLM4TS \cite{chang2023llm4ts}, LLMTime \cite{gruver2024large} and GPT4TS \cite{zhou2023one}.
Additionally, research works like PromptCast \cite{xue2023promptcast}, UniTime \cite{liu2024unitime} and TimeLLM \cite{jin2024time} have developed effective prompt strategies to enable LLMs to perform forecasting tasks.

Despite the elaborate architectural designs, achieving accurate forecasting for non-stationary time series remains challenging \cite{kim2021reversible}.
In this paper, instead of focusing solely on architectural improvements, we tackle time series forecasting by addressing the inherent challenges posed by non-stationarity.

\section{Over-smoothing Problem}
\label{sec: over-smoothing}
\begin{figure}[h]
\centering
\includegraphics[width=0.40\textwidth]{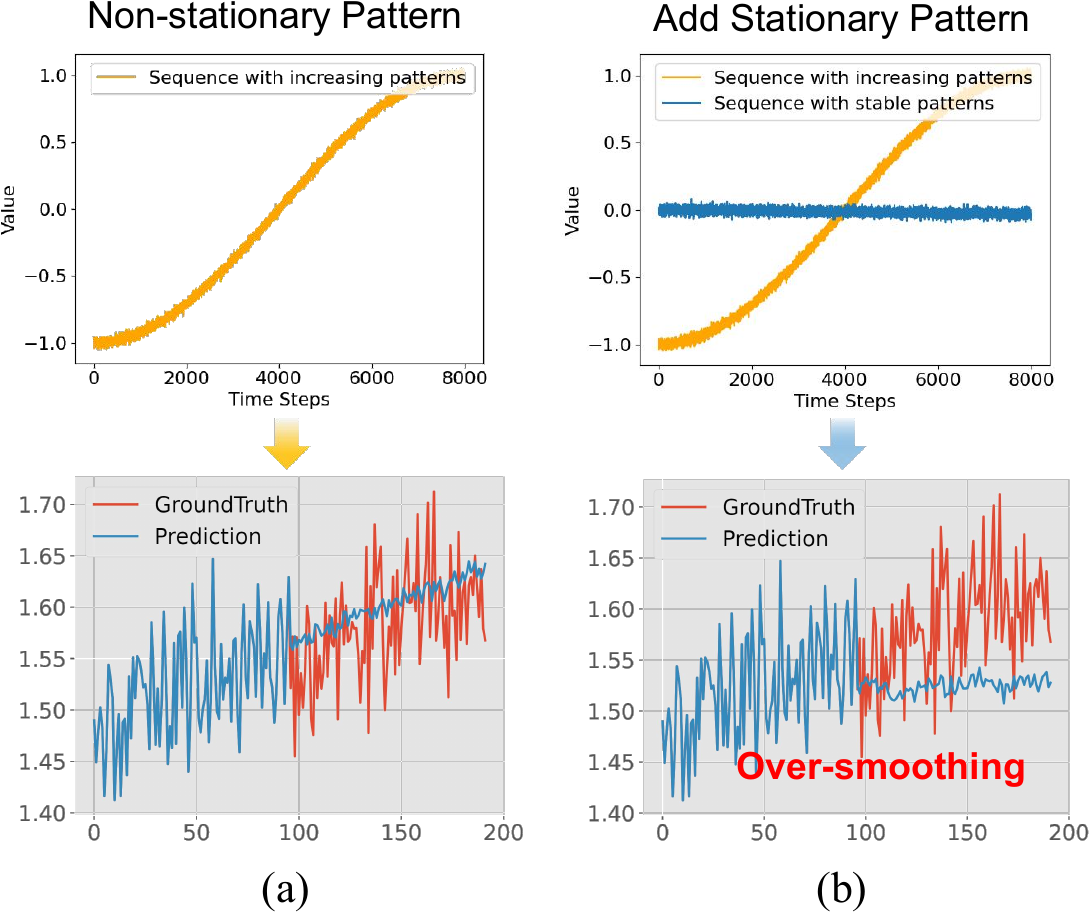} 
\caption{Instance normalization leads to over-smoothed predictions. (a) Predictions using instance normalization when the training set contains only non-stationary patterns.
(b) Predictions using instance normalization when the training set contains both non-stationary and stationary patterns.}
\label{fig: fig10}
\end{figure}

To further explore the over-smoothing problem, we conduct experiments using RLinear on non-stationary sequences with a clear trend and stationary sequences fluctuating around zero.
When the model is trained on non-stationary sequences, it performs well because it can detect and predict the upward trend accurately, as shown in Fig.~\ref{fig: fig10} (a). 
However, when the model is trained on both non-stationary sequences and stationary sequences, it fails to distinguish between the two patterns, resulting in over-smoothed predictions, as shown in Fig.~\ref{fig: fig10} (b).

Although the original sequences are clearly identifiable (e.g., the input fluctuates around zero for stationary sequences), instance normalization aligns all sample distributions and blurs the pattern boundaries. 
As a result, the model struggles to distinguish the underlying non-stationary patterns of samples by the normalized input sequence.
The model tends to learn average predictions to minimize the overall MSE loss, i.e., the over-smoothing problem.

This phenomenon emphasizes the importance of the model leveraging the original data features to distinguish between different patterns, especially for channel independent models that treat samples from different channels equally.

\section{Experimental Setup}
\label{sec: experimental setup}
We follow the standard protocol for data preprocessing. Specifically, we split the datasets into training, validation, and testing by a ratio of 6:2:2 for the ETT datasets and 7:1:2 for the other datasets \cite{liu2024itransformer}.
Following iTransformer \cite{liu2024itransformer}, we used ADAM \cite{diederik2014adam} as the default optimizer across all the experiments. We employed Mean Absolute Error (MAE) and Mean Squared Error (MSE) as evaluation metrics and a lower MSE/MAE value indicates a better performance.
The historical windows length is set to \(L = 96\) for all models and the horizon windows length \(H = \{96, 192, 336, 720\}\). 
All prediction tasks set the drop-last to False. We conducted grid search to optimize the following parameter, i.e., the channel ratio \(\alpha = \{0.1, 0.3, 0.5, 0.7, 1\}\).

\section{Parameter Sensitivity} 
\label{sec: parameter sensitivity}
\subsection{Impact of the Hyperparameter \(\alpha = \lfloor \frac{k}{N} \rfloor\)}
\begin{figure}[h]
    \centering
    \begin{subfigure}{0.22\textwidth}
        \centering
        \includegraphics[width=\linewidth]{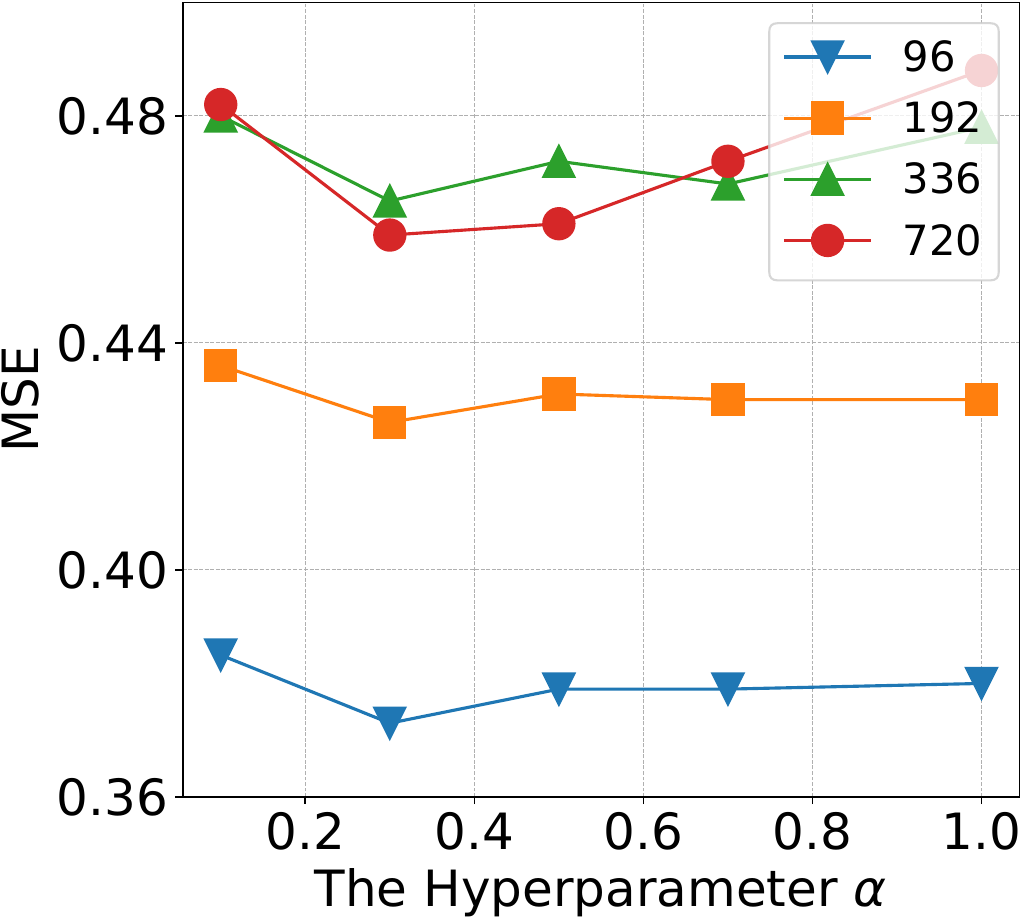}
        \caption{ETTh1}
        \label{fig:fig5-left}
    \end{subfigure}
    \hspace{0.02\textwidth}
    \begin{subfigure}{0.22\textwidth}
        \centering
        \includegraphics[width=\linewidth]{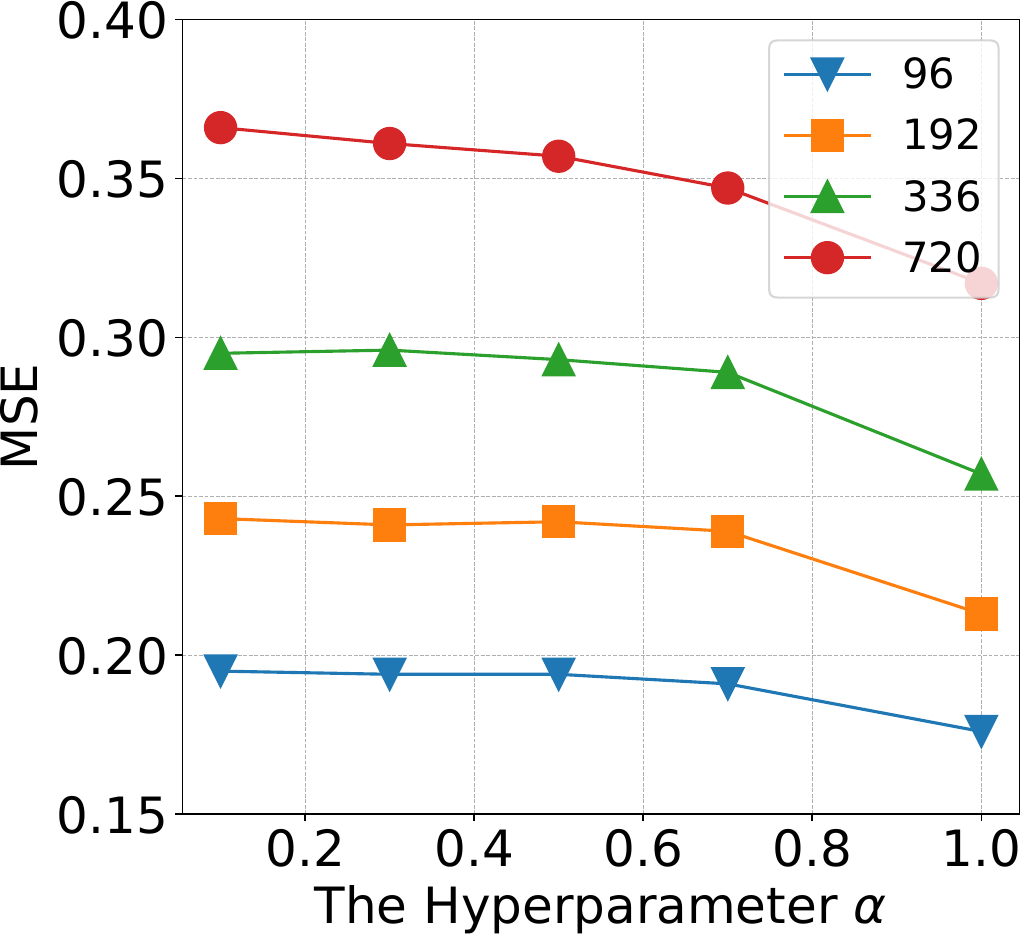}
        \caption{Weather}
        \label{fig:fig5-right}
    \end{subfigure}
    \caption{Performance comparison on the ETTh1 and Weather dataset with varying \(\alpha\).}
    \label{fig: fig5}
\end{figure}

The Channel Selector depends on the hyperparameter \(\alpha\), which determines the ratio of selected channels involved in the fusion process. 
To evaluate the impact of \(\alpha\), we perform a sensitivity analysis on the ETTh1 and Weather datasets, as shown in Fig.~\ref{fig: fig5}. 
Our analysis reveals that the long-term forecasts of CDFM are more sensitive to \(\alpha\) than the short-term forecasts.
Furthermore, the optimal value of \(\alpha\) differs across datasets: for ETTh1, the best results are achieved with \(\alpha=0.3\), whereas for Weather, the optimal configuration is \(\alpha=1\).
This discrepancy can be attributed to the comprehensive influence of non-stationary, similarity, and distributional consistency among channels of different datasets.

\subsection{Increasing Historical Window Length}
The non-stationarity of input sequences is also influenced by the length of the historical window \cite{ye2024frequency}, which in turn affects the performance of deep learning models \cite{liu2023koopa}. 
We evaluate CDFM's forecasting accuracy under varying input lengths \(L=\{48, 96, 192, 336\}\) using the ETTh1 dataset.
As shown in Table~\ref{tab: tab7}, the results indicate consistent improvements in prediction accuracy as the input length increases. 
This is due to the fact that a longer historical window allows the model to better capture long-term dependencies and adapt to non-stationarity of time series.
\begin{table}[h]
\caption{MSE results of CDFM on ETTh1 with varied historical window length \(L\).}
\centering
\large

\renewcommand{\arraystretch}{0.80}
\resizebox{0.38\textwidth}{!}{
\begin{tabular}{c|cccc}
\toprule
Horizon & \begin{tabular}[c]{@{}c@{}}CDFM \\ (\(L\)=48)\end{tabular} & \begin{tabular}[c]{@{}c@{}}CDFM\\  (\(L\)=96)\end{tabular} & \begin{tabular}[c]{@{}c@{}}CDFM\\  (\(L\)=192)\end{tabular} & \begin{tabular}[c]{@{}c@{}}CDFM \\ (\(L\)=336)\end{tabular} \\ \midrule
96     & 0.382                                                  & 0.374                                                   & 0.373                                                   & {\color[HTML]{FF0000}\textbf{0.369}}                                                   \\
192    & 0.437                                                  & 0.426                                                   & 0.415                                                   & {\color[HTML]{FF0000}\textbf{0.403}}                                                  \\
336    & 0.484                                                  & 0.465                                                   & 0.445                                                   & {\color[HTML]{FF0000}\textbf{0.427}}                                                   \\
720    & 0.481                                                  & 0.459                                                   & 0.437                                                   & {\color[HTML]{FF0000}\textbf{0.431}}                                                  \\ \midrule
Avg    & 0.446                                                  & 0.431                                                   & 0.418                                                   &{\color[HTML]{FF0000} \textbf{ 0.408 }}      \\
\bottomrule
\end{tabular}}
\vspace{-2mm}
\label{tab: tab7}
\end{table}

\section{Channel Selection Mechanism}
\label{sec: channel selector}
Take the ETTh2 dataset as an example to illustrate how the channel selector integrates three key factors when selecting channels: non-stationarity, similarity, and distribution consistency.

Table~\ref{tab: channel} presents the detailed metric results for the channels in the ETTh2 dataset, including the non-stationarity and similarity and distribution consistency for each channel. 
'Stationary Loss' denotes the loss of the stationary predictor on the validation set and 'Fusion Loss' denotes the fusion loss between the stationary and non-stationary predictors on the validation set. A significantly higher fusion loss compared to the stationary loss suggests that the channel can exhibit  distribution shifts.

\begin{figure*}[!t]
\centering
\includegraphics[width=0.96\textwidth]{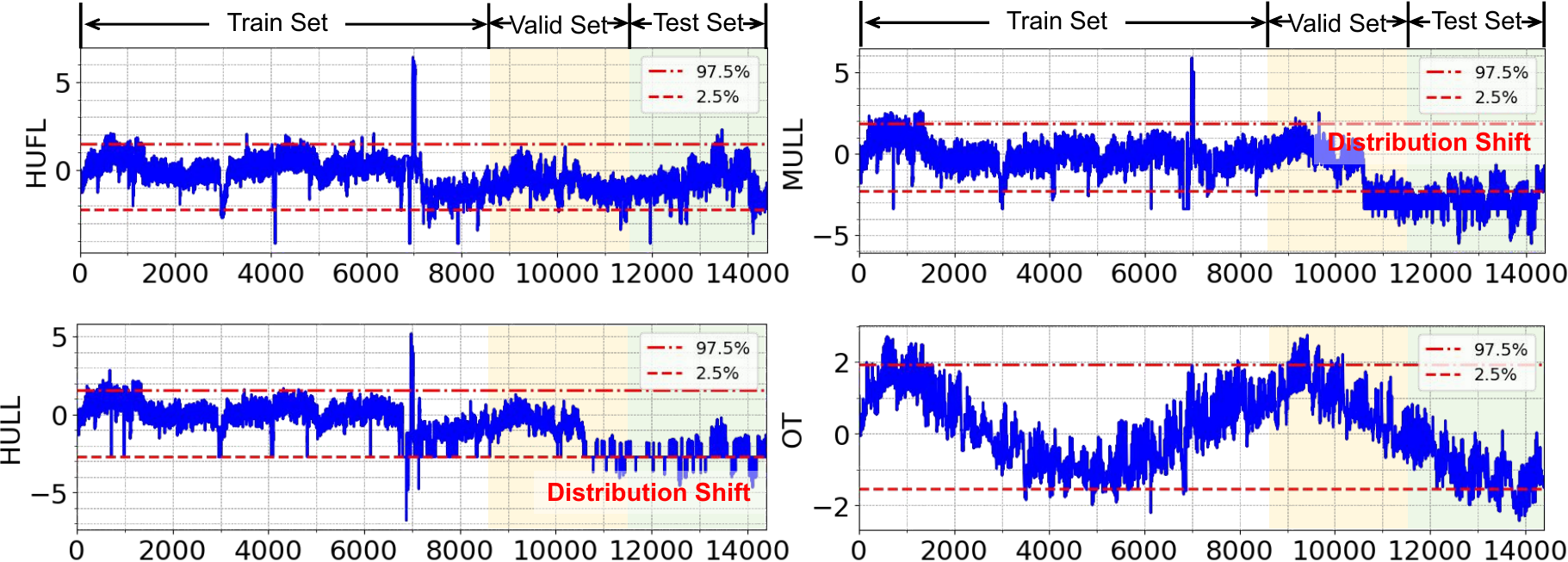} 
\caption{Four channels from the ETTh2 dataset. The red dashed lines are the 97.5\% quantile and 2.5\% quantile of the training set, respectively. Clearly, ‘HULL’ and 'MULL' exhibit distributional shift between the training and test sets.Additionally, the 'OT' channel displays a distinctly different pattern compared to the other channels.}
\label{fig: fig11}
\end{figure*}

\begin{itemize}[left=0em]
\item \textbf{Non-stationarity \& Similarity.} Considering both the non-stationarity and the similarity between channels, we select the top-k channels of the ETTh2 dataset. 
Specifically, we choose the channels 'HUFL', 'HULL', 'MUFL', and 'MULL', where \(k=\lfloor \alpha \times N \rfloor = 4\).
\item \textbf{Distribution Consistency.} Based on the validation set loss, we infer that the 'HULL' and 'MULL' channels exhibit distribution shifts. 
Therefore, both of them are excluded from the final selection of channels to fuse. 
Fig.~\ref{fig: fig11} visually confirms the distribution shifts between the training and test sets for these channels.
\end{itemize}
Consequently, we select the remaining channels, 'HUFL' and 'MUFL', for fusion. The Channel Selection Mechanism enhances predictive accuracy by adapting to the underlying temporal dynamics and avoids the risks of incorporating unreliable non-stationary information, ensuring the effectiveness and robustness of the non-stationary fusion process.

\begin{table}[h]
\caption{The metric results of ETTh2-96-96 task with \(\alpha=0.7\).}
\centering
\large
\renewcommand{\arraystretch}{0.95}
\resizebox{0.46\textwidth}{!}{  
\begin{tabular}{c|c|c|c|c|c}
\toprule
\multirow{3}{*}{Channel} & \multirow{3}{*}{{\begin{tabular}[c]{@{}c@{}}Non-\\ stationarity\end{tabular}}} & \multirow{3}{*}{Similarity} & \multicolumn{3}{c}{Distribution Consistency} \\ \cmidrule{4-6}
                         &                                   &                             & {{\begin{tabular}[c]{@{}c@{}}Stationary\\ Loss\end{tabular}}}    & {{\begin{tabular}[c]{@{}c@{}}Fusion\\ Loss\end{tabular}}}
                     & Consistency \\ \midrule
HUFL                     & 0.569                             & 0.510                       & 0.273            & 0.270       & \(\checkmark\)            \\
HULL                     & 0.590                             & 0.530                       & 0.252            & 0.873       & \(\times\)            \\
MUFL                     & 0.295                             & 0.451                       & 0.082            & 0.083       &  \(\checkmark\)            \\
MULL                     & 0.664                             & 0.395                       & 0.414            & 0.946       & \(\times\)            \\
LUFL                     & 0.290                             & 0.353                       & 0.254            & 0.254       &  \(\checkmark\)            \\
LULL                     & 0.121                             & 0.314                       & 0.013            & 0.013       &  \(\checkmark\)            \\
OT                       & 0.422                             & 0.013                       & 0.183            & 0.183       & \(\checkmark\)          \\ \bottomrule
\end{tabular}}
\label{tab: channel}
\end{table}

\section{Visualization of Forecast Results}
To further demonstrate the forecasting performance of CDFM, we visualize the predictions on ETTm1 with \(\alpha=1\) and ETTh2 with \(\alpha=0.7\), comparing CDFM with DLinear, TimeMixer, NS-Trans, PatchTST, and iTransformer. 

\begin{figure*}[t]
    \centering
    \begin{subfigure}{0.20\textwidth}
        \centering
        \includegraphics[width=\linewidth]{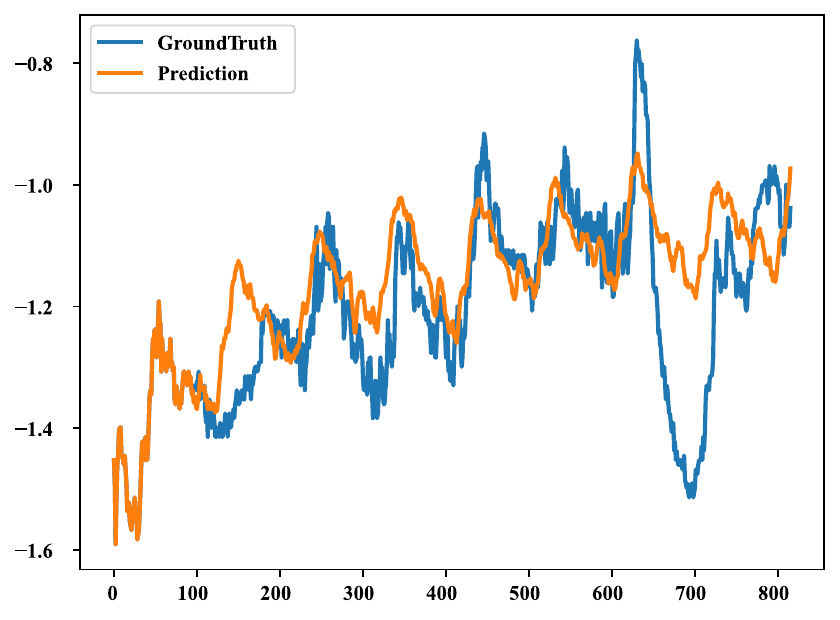}
        \caption{CDFM}
        \label{fig:fig7-1}
    \end{subfigure}
    \hspace{0.02\textwidth} 
    \begin{subfigure}{0.20\textwidth}
        \centering
        \includegraphics[width=\linewidth]{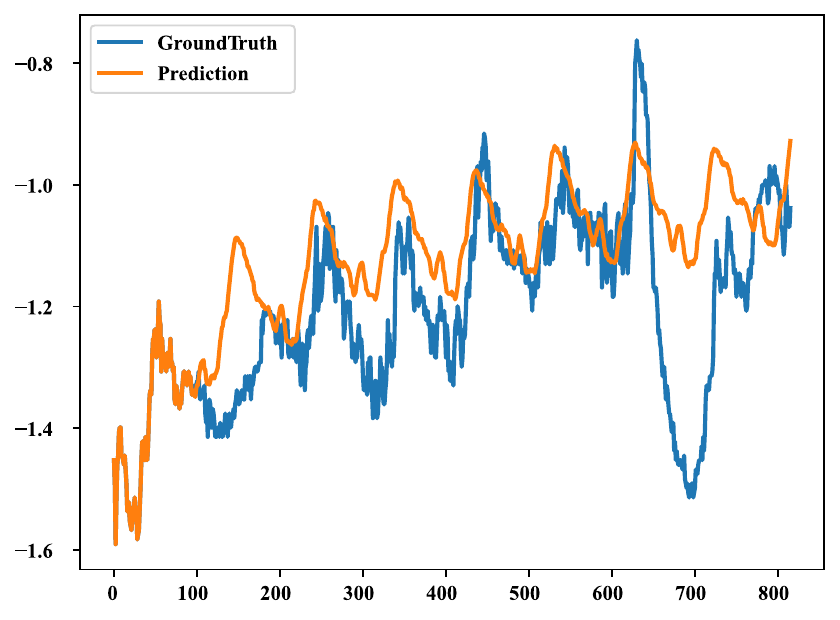}
        \caption{DLinear}
        \label{fig:fig7-2}
    \end{subfigure}
    \hspace{0.02\textwidth} 
    \begin{subfigure}{0.20\textwidth}
        \centering
        \includegraphics[width=\linewidth]{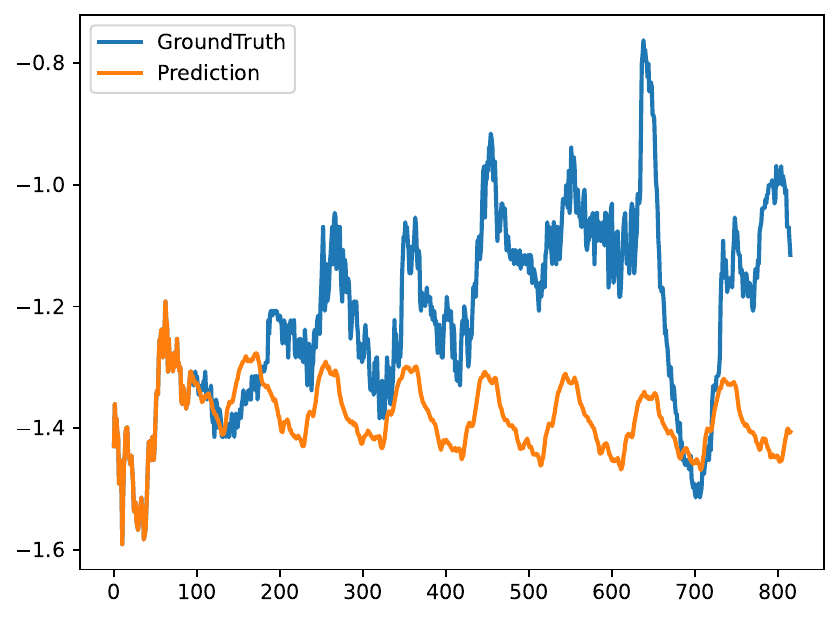}
        \caption{TimeMixer}
        \label{fig:fig7-3}
    \end{subfigure}
    
    \begin{subfigure}{0.20\textwidth}
        \centering
        \includegraphics[width=\linewidth]{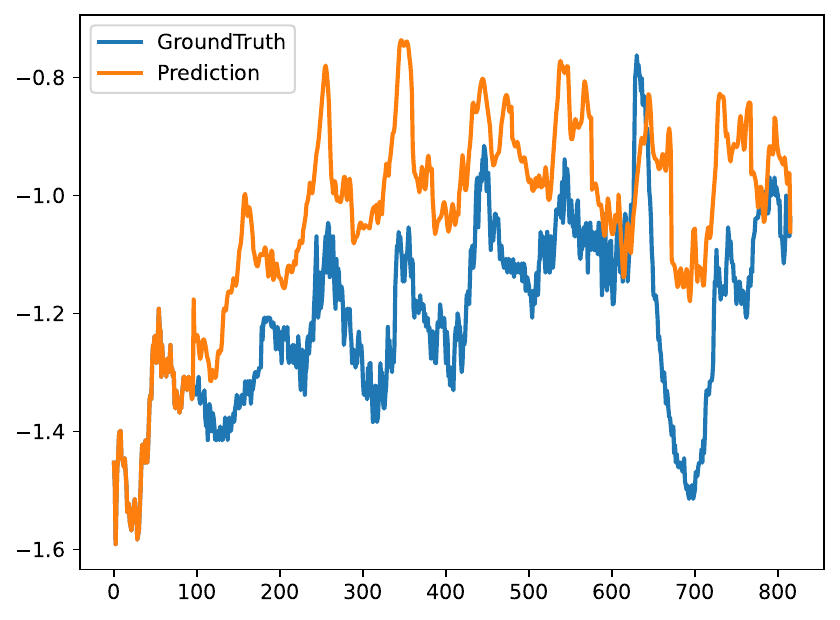}
        \caption{NS-Trans}
        \label{fig:fig7-4}
    \end{subfigure}
    \hspace{0.02\textwidth} 
    \begin{subfigure}{0.20\textwidth}
        \centering
        \includegraphics[width=\linewidth]{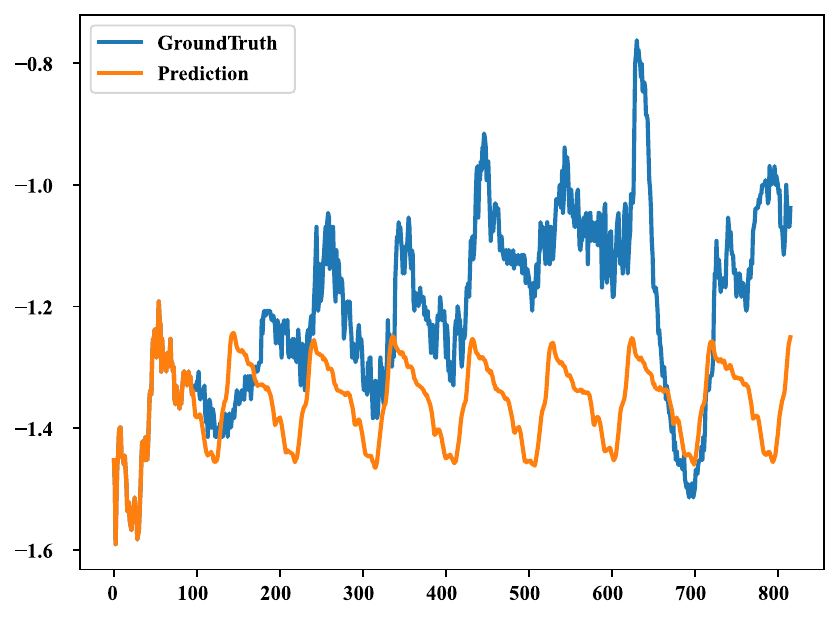}
        \caption{PatchTST}
        \label{fig:fig7-5}
    \end{subfigure}
    \hspace{0.02\textwidth}
    \begin{subfigure}{0.20\textwidth}
        \centering
        \includegraphics[width=\linewidth]{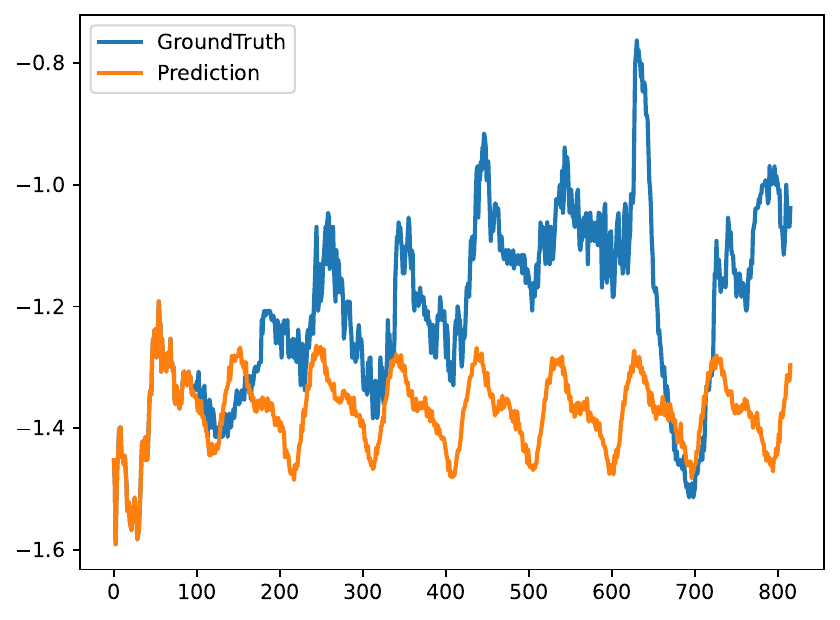}
        \caption{iTransformer}
        \label{fig:fig7-6}
    \end{subfigure}
    \caption{Visualization of input-96-predict-720 results for non-stationary sample from the ETTm1 dataset.}
    \label{fig: fig7}
\end{figure*}

(1) Non-stationary sample (Fig.~\ref{fig: fig7}): For the non-stationary series, we observe that CDFM effectively leverages non-stationary information to capture distribution shifts, achieving the most accurate predictions. 
This highlights CDFM’s strength in modeling complex temporal dependencies and recovering crucial non-stationary information. In contrast, models like TimeMixer, PatchTST, and iTransformer, which rely heavily on normalization, produce over-smoothed predictions and fail to capture the upward trend in the series.

\begin{figure*}[t]
    \centering
    \begin{subfigure}{0.20\textwidth}
        \centering
        \includegraphics[width=\linewidth]{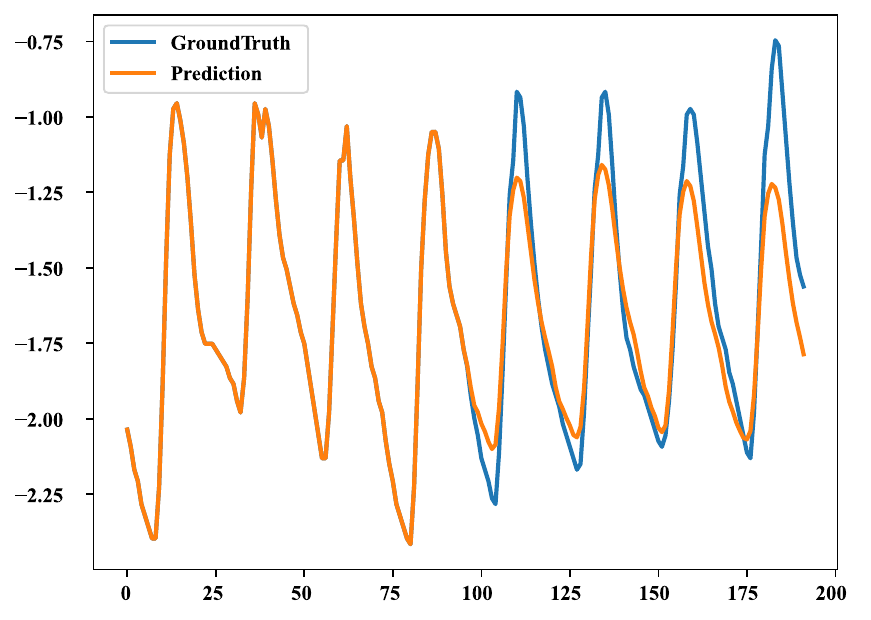}
        \caption{CDFM}
        \label{fig:fig8-1}
    \end{subfigure}
    \hspace{0.02\textwidth} 
    \begin{subfigure}{0.20\textwidth}
        \centering
        \includegraphics[width=\linewidth]{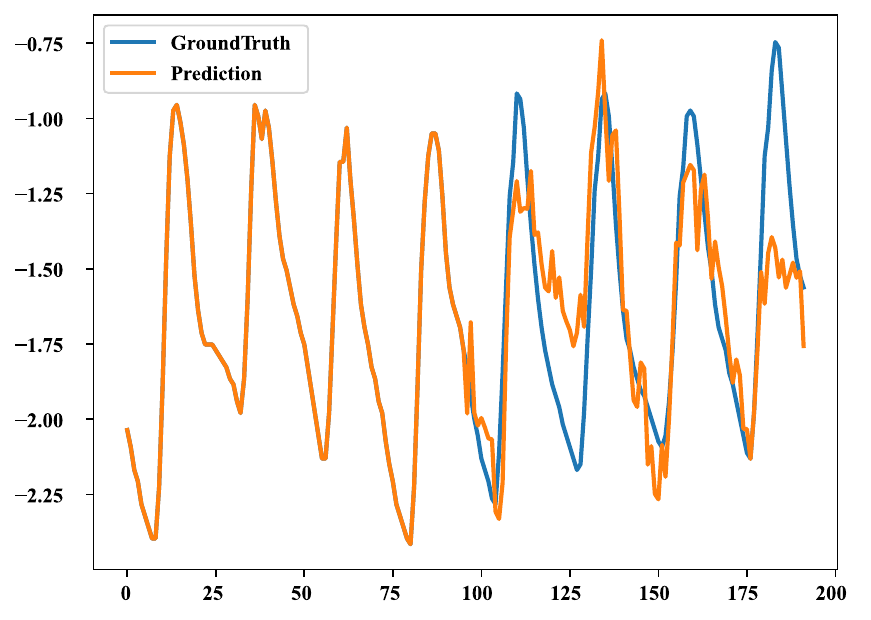}
        \caption{DLinear}
        \label{fig:fig8-2}
    \end{subfigure}
    \hspace{0.02\textwidth}
    \begin{subfigure}{0.20\textwidth}
        \centering
        \includegraphics[width=\linewidth]{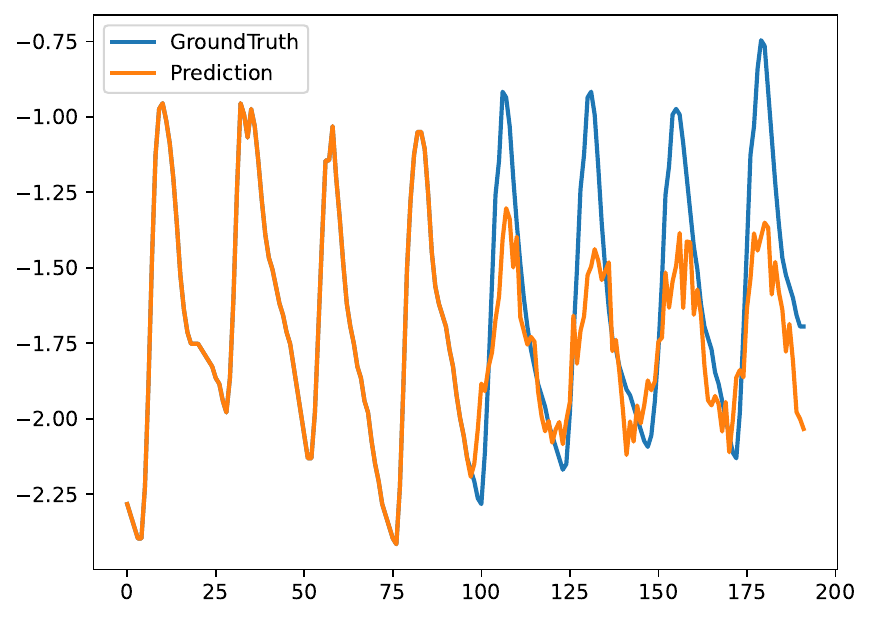}
        \caption{TimeMixer}
        \label{fig:fig8-3}
    \end{subfigure}
    
    \begin{subfigure}{0.20\textwidth}
        \centering
        \includegraphics[width=\linewidth]{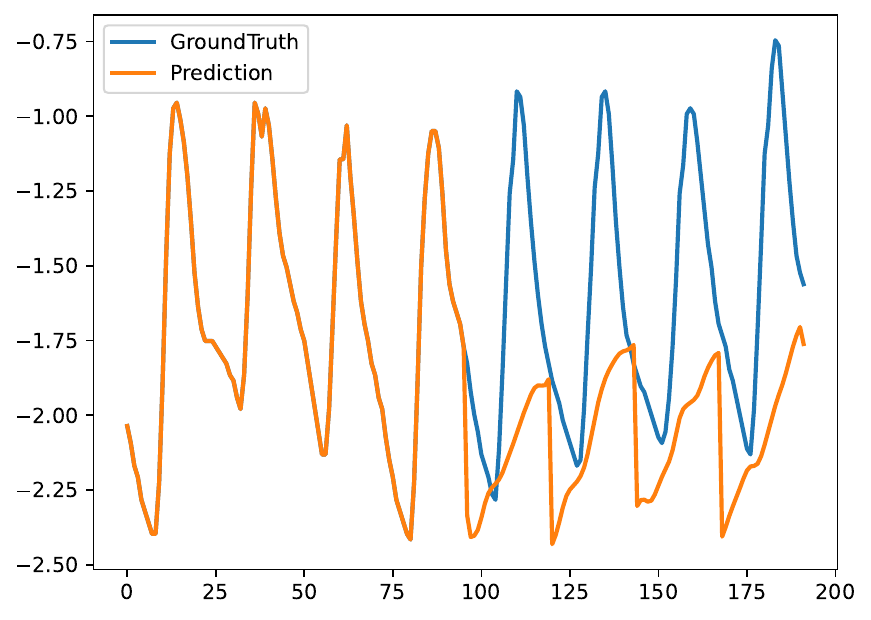}
        \caption{NS-Trans}
        \label{fig:fig8-4}
    \end{subfigure}
    \hspace{0.02\textwidth}
    \begin{subfigure}{0.20\textwidth}
        \centering
        \includegraphics[width=\linewidth]{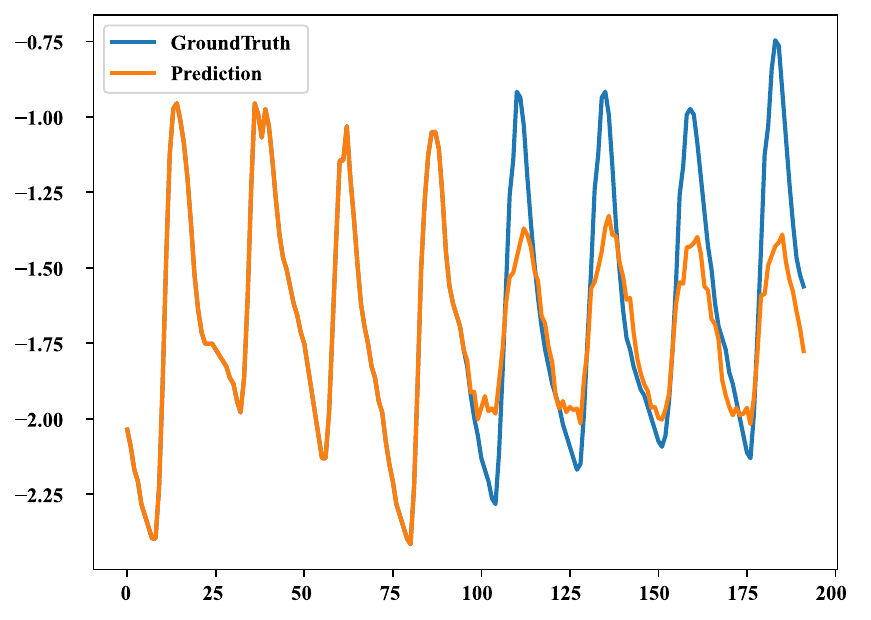}
        \caption{PatchTST}
        \label{fig:fig8-5}
    \end{subfigure}
    \hspace{0.02\textwidth} 
    \begin{subfigure}{0.20\textwidth}
        \centering
        \includegraphics[width=\linewidth]{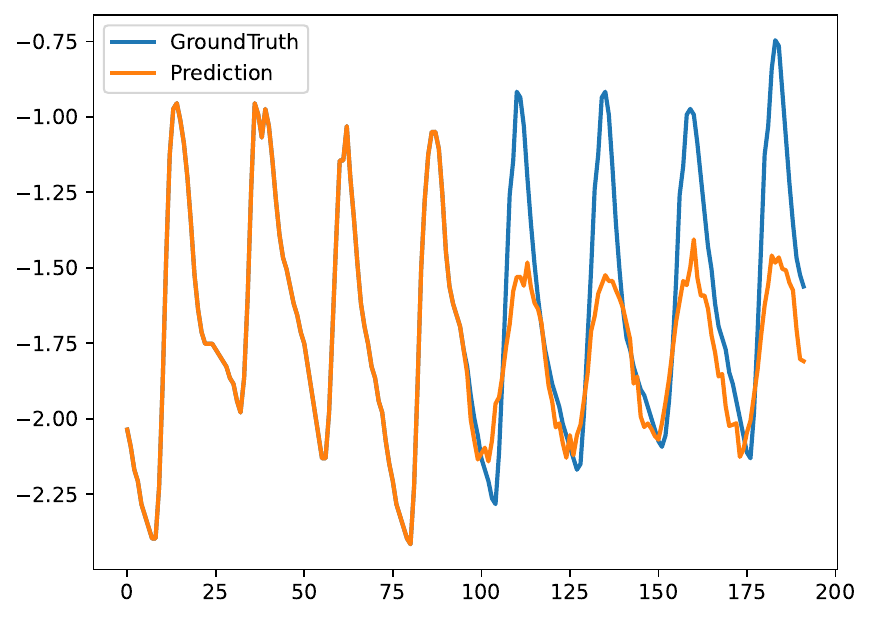}
        \caption{iTransformer}
        \label{fig:fig8-6}
    \end{subfigure}

    \caption{Visualization of input-96-predict-96 results for relatively stationary sample from the ETTh2 dataset.}
    \label{fig: fig8}
\end{figure*}

(2) Relatively stationary sample (Fig.~\ref{fig: fig8}): For the relatively stationary series, CDFM maintains its predictability without being adversely affected by non-stationary information. This is attributed to the dynamic fusion mechanism, which enables CDFM to preserve stable patterns in the data, ensuring that the model accurately captures stationary components. Unlike NS-Trans, which is overly sensitive to non-stationary fluctuations, CDFM remains robust and effective.

These results underscore the effectiveness of CDFM in handling both stationary and non-stationary time 
series, and the dynamic fusion mechanism enables CDFM to adapt to varying data characteristics. 
Importantly, CDFM avoids the over-smoothing problem seen in models that rely exclusively on normalization.

\vfill

\section{Full Results}
\label{sec: full results}
We thoroughly evaluate the proposed framework CDFM on a variety of time series forecasting tasks, demonstrating its robust performance across different datasets and forecasting horizons. 
Due to space limitations in the main text, the full prediction results are provided in the following section.

Since the historical length \(L\) is a crucial hyperparameter that not only determines the richness of past information the model can leverage but also determines the number of parameters of the model, particularly for linear models.
We also conduct multivariate time forecasting experiments on searched hyperparameter input lengths. Specifically, we search for input length among \(L=\{96, 336, 512\}\) and the full results are presented in Table~\ref{tab: tab2_mul_input}.
We can observe that CDFM achieves consistent SOTA performance in all benchmarks on non-stationary datasets. 
While PatchTST shows competitive results across several datasets, CDFM still outperforms it, with an average reduction of 1.4\% in MSE and 1.2\% in MAE.
Furthermore, CDFM provides a well-rounded solution that balances accuracy and efficiency effectively, achieving superior performance with fewer parameters.

Table~\ref{tab: tab3_all} contains the detailed results compared to existing instance normalization techniques.
Table~\ref{tab: tab4_all} includes detailed results compared to non-stationary models.
And Table~\ref{tab: tab5_all} contains the detailed performance promotion obtained by applying our fuse framework.

\begin{table*}[h]
\caption{Multivariate forecasting results with prediction lengths \(H \in
\{96, 192, 336, 720\}\) and best input length \(L = \{96, 336, 512\}\) and baseline results are from DUET \cite{qiu2024duet}. Best results are highlighted in {\color[HTML]{FF0000} \textbf{red}} and the second {\textcolor{blue} {\ul underlined}}.}
\centering
\large
\renewcommand{\arraystretch}{0.80}
\resizebox{0.87\textwidth}{!}{
\begin{tabular}{c|c|cc|cc|cc|cc|cc|cc|cc}
\toprule
\multicolumn{2}{c|}{Models}             & \multicolumn{2}{c|}{\begin{tabular}[c]{@{}c@{}}CDFM\\ (ours)\end{tabular}}     & \multicolumn{2}{c|}{\begin{tabular}[c]{@{}c@{}}iTransformer\\ (2024)\end{tabular}} & \multicolumn{2}{c|}{\begin{tabular}[c]{@{}c@{}}TimeMixer\\ (2024)\end{tabular}} & \multicolumn{2}{c|}{\begin{tabular}[c]{@{}c@{}}FITS\\ (2024)\end{tabular}}     & \multicolumn{2}{c|}{\begin{tabular}[c]{@{}c@{}}PatchTST\\ (2023)\end{tabular}} & \multicolumn{2}{c|}{\begin{tabular}[c]{@{}c@{}}TimesNet\\ (2023a)\end{tabular}} & \multicolumn{2}{c}{\begin{tabular}[c]{@{}c@{}}DLinear\\ (2023)\end{tabular}}  \\ \midrule
\multicolumn{2}{c|}{Metric}             & MSE                                   & MAE                                   & MSE                                     & MAE                                     & MSE                                    & MAE                                   & MSE                                   & MAE                                   & MSE                                   & MAE                                   & MSE                                    & MAE                                   & MSE                                   & MAE                                   \\ \midrule
                                 & 96  & {\color[HTML]{FF0000} \textbf{0.368}} & {\color[HTML]{FF0000} \textbf{0.393}} & 0.386                                   & 0.405                                   & {\textcolor{blue} {\ul 0.372}}     & 0.401                                 & 0.376                                 & {\textcolor{blue} {\ul 0.396}}    & 0.377                                 & 0.397                                 & 0.389                                  & 0.412                                 & 0.379                                 & 0.403                                 \\
                                 & 192 & {\textcolor{blue} {\ul 0.402}}    & {\color[HTML]{FF0000} \textbf{0.413}} & 0.424                                   & 0.440                                   & 0.413                                  & 0.430                                 & {\color[HTML]{FF0000} \textbf{0.400}} & {\textcolor{blue} {\ul 0.418}}    & 0.409                                 & 0.425                                 & 0.440                                  & 0.443                                 & 0.408                                 & 0.419                                 \\
                                 & 336 & {\color[HTML]{FF0000} \textbf{0.418}} & {\color[HTML]{FF0000} \textbf{0.432}} & 0.449                                   & 0.460                                   & 0.438                                  & 0.450                                 & {\textcolor{blue} {\ul 0.419}}    & {\textcolor{blue} {\ul 0.435}}    & 0.431                                 & 0.444                                 & 0.523                                  & 0.487                                 & 0.440                                 & 0.440                                 \\
                                 & 720 & {\color[HTML]{FF0000} \textbf{0.431}} & {\color[HTML]{FF0000} \textbf{0.449}} & 0.495                                   & 0.487                                   & 0.486                                  & 0.484                                 & {\textcolor{blue} {\ul 0.435}}    & {\textcolor{blue} {\ul 0.458}}    & 0.457                                 & 0.477                                 & 0.521                                  & 0.495                                 & 0.471                                 & 0.493                                 \\ \cmidrule(rl){2-16}
\multirow{-5}{*}{\rotatebox{90}{ETTh1}}          & Avg & {\color[HTML]{FF0000} \textbf{0.405}} & {\color[HTML]{FF0000} \textbf{0.422}} & 0.439                                   & 0.448                                   & 0.427                                  & 0.441                                 & {\textcolor{blue} {\ul 0.408}}    & {\textcolor{blue} {\ul 0.427}}    & 0.419                                 & 0.436                                 & 0.468                                  & 0.459                                 & 0.425                                 & 0.439                                 \\ \midrule
                                 & 96  & {\color[HTML]{FF0000} \textbf{0.269}} & {\color[HTML]{FF0000} \textbf{0.335}} & 0.297                                   & 0.348                                   & 0.281                                  & 0.351                                 & 0.277                                 & 0.345                                 & {\textcolor{blue} {\ul 0.274}}    & {\textcolor{blue} {\ul 0.337}}    & 0.334                                  & 0.370                                 & 0.300                                 & 0.364                                 \\
                                 & 192 & {\textcolor{blue} {\ul 0.335}}    & {\color[HTML]{FF0000} \textbf{0.378}} & 0.372                                   & 0.403                                   & 0.349                                  & 0.387                                 & {\color[HTML]{FF0000} \textbf{0.331}} & {\textcolor{blue} {\ul 0.379}}    & 0.348                                 & 0.384                                 & 0.404                                  & 0.413                                 & 0.387                                 & 0.423                                 \\
                                 & 336 & {\textcolor{blue} {\ul 0.362}}    & {\textcolor{blue} {\ul 0.402}}    & 0.388                                   & 0.417                                   & 0.366                                  & 0.413                                 & {\color[HTML]{FF0000} \textbf{0.350}} & {\color[HTML]{FF0000} \textbf{0.396}} & 0.377                                 & 0.416                                 & 0.389                                  & 0.435                                 & 0.490                                 & 0.487                                 \\
                                 & 720 & {\textcolor{blue} {\ul 0.389}}    & {\color[HTML]{FF0000} \textbf{0.416}} & 0.424                                   & 0.444                                   & 0.401                                  & 0.436                                 & {\color[HTML]{FF0000} \textbf{0.382}} & 0.425                                 & 0.406                                 & 0.441                                 & 0.434                                  & 0.448                                 & 0.704                                 & 0.597                                 \\ \cmidrule(rl){2-16}
\multirow{-5}{*}{\rotatebox{90}{ETTh2}}          & Avg & {\textcolor{blue} {\ul 0.339}}    & {\color[HTML]{FF0000} \textbf{0.383}} & 0.370                                   & 0.403                                   & 0.349                                  & 0.397                                 & 0.335                                 & 0.386                                 & 0.351                                 & 0.395                                 & 0.390                                  & 0.417                                 & 0.470                                 & 0.468                                 \\ \midrule
                                 & 96  & 0.297                                 & {\color[HTML]{FF0000} \textbf{0.341}} & 0.300                                   & 0.353                                   & {\textcolor{blue} {\ul 0.293}}     & 0.345                                 & 0.303                                 & 0.345                                 & {\color[HTML]{FF0000} \textbf{0.289}} & {\textcolor{blue} {\ul 0.343}}    & 0.340                                  & 0.378                                 & 0.300                                 & 0.345                                 \\
                                 & 192 & {\color[HTML]{FF0000} \textbf{0.327}} & {\color[HTML]{FF0000} \textbf{0.360}} & 0.341                                   & 0.380                                   & 0.335                                  & 0.372                                 & 0.337                                 & {\textcolor{blue} {\ul 0.365}}    & {\textcolor{blue} {\ul 0.329}}    & 0.368                                 & 0.392                                  & 0.404                                 & 0.336                                 & 0.366                                 \\
                                 & 336 & {\textcolor{blue} {\ul 0.363}}    & {\color[HTML]{FF0000} \textbf{0.381}} & 0.374                                   & 0.396                                   & 0.368                                  & 0.386                                 & 0.368                                 & {\textcolor{blue} {\ul 0.384}}    & {\color[HTML]{FF0000} \textbf{0.362}} & 0.390                                 & 0.423                                  & 0.426                                 & 0.367                                 & 0.386                                 \\
                                 & 720 & 0.426                                 & 0.417                                 & 0.429                                   & 0.430                                   & 0.426                                  & 0.417                                 & {\textcolor{blue} {\ul 0.420}}    & {\color[HTML]{FF0000} \textbf{0.413}} & {\color[HTML]{FF0000} \textbf{0.416}} & 0.423                                 & 0.475                                  & 0.453                                 & 0.419                                 & {\textcolor{blue} {\ul 0.416}}    \\ \cmidrule(rl){2-16}
\multirow{-5}{*}{\rotatebox{90}{ETTm1}}          & Avg & {\textcolor{blue} {\ul 0.353}}    & {\color[HTML]{FF0000} \textbf{0.375}} & 0.361                                   & 0.390                                   & 0.356                                  & 0.380                                 & 0.357                                 & {\textcolor{blue} {\ul 0.377}}    & {\color[HTML]{FF0000} \textbf{0.349}} & 0.381                                 & 0.408                                  & 0.415                                 & 0.356                                 & 0.378                                 \\ \midrule
                                 & 96  & {\textcolor{blue} {\ul 0.165}}    & {\color[HTML]{FF0000} \textbf{0.254}} & 0.175                                   & 0.266                                   & {\textcolor{blue} {\ul 0.165}}     & 0.256                                 & {\textcolor{blue} {\ul 0.165}}    & {\color[HTML]{FF0000} \textbf{0.254}} & {\textcolor{blue} {\ul 0.165}}    & 0.255                                 & 0.189                                  & 0.265                                 & {\color[HTML]{FF0000} \textbf{0.164}} & 0.255                                 \\
                                 & 192 & {\textcolor{blue} {\ul 0.220}}    & {\color[HTML]{FF0000} \textbf{0.291}} & 0.242                                   & 0.312                                   & 0.225                                  & 0.298                                 & {\color[HTML]{FF0000} \textbf{0.219}} & {\color[HTML]{FF0000} \textbf{0.291}} & 0.221                                 & 0.293                                 & 0.254                                  & 0.310                                 & 0.224                                 & 0.304                                 \\
                                 & 336 & {\color[HTML]{FF0000} \textbf{0.271}} & {\color[HTML]{FF0000} \textbf{0.324}} & 0.282                                   & 0.337                                   & 0.277                                  & 0.332                                 & {\textcolor{blue} {\ul 0.272}}    & {\textcolor{blue} {\ul 0.326}}    & 0.276                                 & 0.327                                 & 0.313                                  & 0.345                                 & 0.277                                 & 0.337                                 \\
                                 & 720 & 0.362                                 & {\color[HTML]{FF0000} \textbf{0.380}} & 0.375                                   & 0.394                                   & {\textcolor{blue} {\ul 0.360}}     & 0.387                                 & {\color[HTML]{FF0000} \textbf{0.359}} & {\textcolor{blue} {\ul 0.381}}    & 0.362                                 & 0.381                                 & 0.413                                  & 0.402                                 & 0.371                                 & 0.401                                 \\ \cmidrule(rl){2-16}
\multirow{-5}{*}{\rotatebox{90}{ETTm2}}          & Avg & {\textcolor{blue} {\ul 0.255}}    & {\color[HTML]{FF0000} \textbf{0.312}} & 0.269                                   & 0.327                                   & 0.257                                  & 0.318                                 & {\color[HTML]{FF0000} \textbf{0.254}} & {\textcolor{blue} {\ul 0.313}}    & 0.256                                 & 0.314                                 & 0.292                                  & 0.331                                 & 0.259                                 & 0.324                                 \\ \midrule
                                 & 96  & 0.168                                 & 0.222                                 & 0.157                                   & 0.207                                   & {\color[HTML]{FF0000} \textbf{0.147}}  & {\textcolor{blue} {\ul 0.198}}    & 0.172                                 & 0.225                                 & {\textcolor{blue} {\ul 0.149}}    & {\color[HTML]{FF0000} \textbf{0.196}} & 0.168                                  & 0.214                                 & 0.170                                 & 0.230                                 \\
                                 & 192 & 0.210                                 & 0.257                                 & 0.200                                   & 0.248                                   & {\textcolor{blue} {\ul 0.192}}     & {\textcolor{blue} {\ul 0.243}}    & 0.215                                 & 0.261                                 & {\color[HTML]{FF0000} \textbf{0.191}} & {\color[HTML]{FF0000} \textbf{0.239}} & 0.219                                  & 0.262                                 & 0.216                                 & 0.273                                 \\
                                 & 336 & 0.260                                 & 0.294                                 & 0.252                                   & 0.287                                   & {\textcolor{blue} {\ul 0.247}}     & {\textcolor{blue} {\ul 0.284}}    & 0.261                                 & 0.295                                 & {\color[HTML]{FF0000} \textbf{0.242}} & {\color[HTML]{FF0000} \textbf{0.279}} & 0.278                                  & 0.302                                 & 0.258                                 & 0.307                                 \\
                                 & 720 & {\textcolor{blue} {\ul 0.317}}    & 0.339                                 & 0.320                                   & {\textcolor{blue} {\ul 0.336}}      & 0.318                                  & {\color[HTML]{FF0000} \textbf{0.330}} & 0.326                                 & 0.341                                 & {\color[HTML]{FF0000} \textbf{0.312}} & {\color[HTML]{FF0000} \textbf{0.330}} & 0.353                                  & 0.351                                 & 0.323                                 & 0.362                                 \\ \cmidrule(rl){2-16}
\multirow{-5}{*}{\rotatebox{90}{Weather}}        & Avg & 0.239                                 & 0.278                                 & 0.232                                   & 0.270                                   & {\textcolor{blue} {\ul 0.226}}     & {\textcolor{blue} {\ul 0.264}}    & 0.244                                 & 0.281                                 & {\color[HTML]{FF0000} \textbf{0.224}} & {\color[HTML]{FF0000} \textbf{0.261}} & 0.255                                  & 0.282                                 & 0.242                                 & 0.293                                 \\ \midrule
                                 & 96  & {\color[HTML]{FF0000} \textbf{0.079}} & {\color[HTML]{FF0000} \textbf{0.196}} & 0.086                                   & 0.205                                   & 0.084                                  & 0.207                                 & 0.082                                 & {\textcolor{blue} {\ul 0.199}}    & {\color[HTML]{FF0000} \textbf{0.079}} & 0.200                                 & 0.112                                  & 0.242                                 & {\textcolor{blue} {\ul 0.080}}    & 0.202                                 \\
                                 & 192 & {\color[HTML]{FF0000} \textbf{0.156}} & {\textcolor{blue} {\ul 0.291}}    & 0.177                                   & 0.299                                   & 0.178                                  & 0.300                                 & 0.173                                 & 0.295                                 & {\textcolor{blue} {\ul 0.159}}    & {\color[HTML]{FF0000} \textbf{0.289}} & 0.209                                  & 0.334                                 & 0.182                                 & 0.321                                 \\
                                 & 336 & {\color[HTML]{FF0000} \textbf{0.268}} & {\color[HTML]{FF0000} \textbf{0.389}} & 0.331                                   & 0.417                                   & 0.376                                  & 0.451                                 & 0.317                                 & 0.406                                 & {\textcolor{blue} {\ul 0.297}}    & {\textcolor{blue} {\ul 0.399}}    & 0.358                                  & 0.435                                 & 0.327                                 & 0.434                                 \\
                                 & 720 & {\textcolor{blue} {\ul 0.653}}    & {\textcolor{blue} {\ul 0.609}}    & 0.846                                   & 0.693                                   & 0.884                                  & 0.707                                 & 0.825                                 & 0.684                                 & 0.751                                 & 0.650                                 & 0.944                                  & 0.736                                 & {\color[HTML]{FF0000} \textbf{0.578}} & {\color[HTML]{FF0000} \textbf{0.605}} \\ \cmidrule(rl){2-16}
\multirow{-5}{*}{\rotatebox{90}{Exchange}} & Avg & {\color[HTML]{FF0000} \textbf{0.289}} & {\color[HTML]{FF0000} \textbf{0.371}} & 0.360                                   & 0.404                                   & 0.381                                  & 0.416                                 & 0.349                                 & 0.396                                 & 0.322                                 & {\textcolor{blue} {\ul 0.385}}    & 0.406                                  & 0.437                                 & {\textcolor{blue} {\ul 0.292}}    & 0.391                                 \\ \midrule
                                 & 96  & {\color[HTML]{FF0000} \textbf{0.134}} & {\color[HTML]{FF0000} \textbf{0.229}} & {\color[HTML]{FF0000} \textbf{0.134}}   & {\textcolor{blue} {\ul 0.230}}      & 0.153                                  & 0.256                                 & {\textcolor{blue} {\ul 0.139}}    & 0.237                                 & 0.143                                 & 0.247                                 & 0.169                                  & 0.271                                 & 0.140                                 & 0.237                                 \\
                                 & 192 & {\color[HTML]{FF0000} \textbf{0.149}} & {\color[HTML]{FF0000} \textbf{0.244}} & {\textcolor{blue} {\ul 0.154}}      & {\textcolor{blue} {\ul 0.250}}      & 0.168                                  & 0.269                                 & {\textcolor{blue} {\ul 0.154}}    & {\textcolor{blue} {\ul 0.250}}    & 0.158                                 & 0.260                                 & 0.180                                  & 0.280                                 & {\textcolor{blue} {\ul 0.154}}    & 0.251                                 \\
                                 & 336 & {\color[HTML]{FF0000} \textbf{0.164}} & {\color[HTML]{FF0000} \textbf{0.261}} & 0.169                                   & {\textcolor{blue} {\ul 0.265}}      & 0.189                                  & 0.291                                 & 0.170                                 & 0.268                                 & {\textcolor{blue} {\ul 0.168}}    & 0.267                                 & 0.204                                  & 0.304                                 & 0.169                                 & 0.268                                 \\
                                 & 720 & {\textcolor{blue} {\ul 0.199}}    & {\textcolor{blue} {\ul 0.293}}    & {\color[HTML]{FF0000} \textbf{0.194}}   & {\color[HTML]{FF0000} \textbf{0.288}}   & 0.228                                  & 0.320                                 & 0.212                                 & 0.304                                 & 0.214                                 & 0.307                                 & 0.205                                  & 0.304                                 & 0.204                                 & 0.301                                 \\ \cmidrule(rl){2-16}
\multirow{-5}{*}{\rotatebox{90}{Electricity}}    & Avg & {\color[HTML]{FF0000} \textbf{0.162}} & {\color[HTML]{FF0000} \textbf{0.257}} & {\textcolor{blue} {\ul 0.163}}      & {\textcolor{blue} {\ul 0.258}}      & 0.185                                  & 0.284                                 & 0.169                                 & 0.265                                 & 0.171                                 & 0.270                                 & 0.190                                  & 0.290                                 & 0.167                                 & 0.264                                 \\
\midrule
\rowcolor{pink!50}
\multicolumn{2}{c}{1st Count}          & {\color[HTML]{FF0000} \textbf{15}}    & {\color[HTML]{FF0000} \textbf{25}}    & 2                                       & 1                                       & 1                                      & 1                                     & 7                                     & 4                                     & {\textcolor{blue} {\ul 9}}        & {\textcolor{blue} {\ul 6}}        & 0                                      & 0                                     & 2                                     & 1                                    
\\ \bottomrule  
\end{tabular}}
\label{tab: tab2_mul_input}
\end{table*}

\begin{table*}[h]
\caption{Comparison between CDFM and normalization approaches using DLinear as backbone model with different prediction lengths \(H \in
\{96, 192, 336, 720\}\) and the input length \(L = 96\). Best results are highlighted in {\color[HTML]{FF0000} \textbf{red}} and the second {\textcolor{blue} {\ul underlined}}.}
\centering
\large

\renewcommand{\arraystretch}{0.9}
\resizebox{0.85\textwidth}{!}{
\begin{tabular}{c|c|cc|cc|cc|cc|cc|cc|cc}
\toprule
\multicolumn{2}{c|}{Models}    & \multicolumn{2}{c|}{DLinear} & \multicolumn{2}{c|}{+CDFM}                                                      & \multicolumn{2}{c|}{+RevIN}                                               & \multicolumn{2}{c|}{+Dish-TS} & \multicolumn{2}{c|}{+FAN} & \multicolumn{2}{c|}{+SAN}                    & \multicolumn{2}{c}{+SIN}                    \\ \midrule
\multicolumn{2}{c|}{Metric}    & MSE          & MAE          & MSE                                   & MAE                                   & MSE                                & MAE                                & MSE          & MAE          & MSE        & MAE        & MSE                                & MAE   & MSE                                & MAE   \\ \midrule
                        & 96  & 0.398        & 0.410        & {\color[HTML]{FF0000} \textbf{0.374}} & {\color[HTML]{FF0000} \textbf{0.389}} & 0.383                              & {\textcolor{blue} {\ul 0.391}} & 0.389        & 0.399        & 0.390      & 0.408      & 0.387                              & 0.400 & {\textcolor{blue} {\ul 0.382}} & 0.394 \\
                        & 192 & 0.434        & 0.427        & {\color[HTML]{FF0000} \textbf{0.426}} & {\color[HTML]{FF0000} \textbf{0.419}} & {\textcolor{blue} {\ul 0.434}}                              & {\textcolor{blue} {\ul 0.420}} & 0.443        & 0.433        & 0.439      & 0.436      & 0.436                              & 0.431 & {\textcolor{blue} {\ul 0.434}} & 0.427 \\
                        & 336 & 0.499        & 0.477        & {\color[HTML]{FF0000} \textbf{0.465}} & {\color[HTML]{FF0000} \textbf{0.438}} & {\textcolor{blue} {\ul 0.475}} & {\textcolor{blue} {\ul 0.441}} & 0.487        & 0.456        & 0.497      & 0.469      & 0.488                              & 0.463 & 0.478                              & 0.453 \\
                        & 720 & 0.508        & 0.503        & {\color[HTML]{FF0000} \textbf{0.459}} & {\color[HTML]{FF0000} \textbf{0.455}} & {\textcolor{blue} {\ul 0.475}} & {\textcolor{blue} {\ul 0.465}} & 0.523        & 0.508        & 0.581      & 0.537      & 0.526                              & 0.514 & 0.504                              & 0.500 \\ \cmidrule(rl){2-16}
\multirow{-5}{*}{\rotatebox{90}{ETTh1}} & Avg & 0.460        & 0.454        & {\color[HTML]{FF0000} \textbf{0.431}} & {\color[HTML]{FF0000} \textbf{0.425}} & {\textcolor{blue} {\ul 0.442}} & {\textcolor{blue} {\ul 0.429}} & 0.461        & 0.449        & 0.477      & 0.463      & 0.459                              & 0.452 & 0.450                              & 0.444 \\ \midrule
                        & 96  & 0.346        & 0.374        & {\color[HTML]{FF0000} \textbf{0.324}} & {\color[HTML]{FF0000} \textbf{0.357}} & 0.352                              & {\textcolor{blue} {\ul 0.369}} & 0.343        & 0.375        & 0.337      & 0.374      & {\textcolor{blue} {\ul 0.333}} & 0.374 & 0.350                              & 0.383 \\
                        & 192 & 0.382        & 0.392        & {\color[HTML]{FF0000} \textbf{0.367}} & {\color[HTML]{FF0000} \textbf{0.379}} & 0.388                              & {\textcolor{blue} {\ul 0.386}} & 0.381        & 0.391        & 0.376      & 0.394      & {\textcolor{blue} {\ul 0.374}} & 0.396 & 0.383                              & 0.396 \\
                        & 336 & 0.414        & 0.414        & {\color[HTML]{FF0000} \textbf{0.398}} & {\color[HTML]{FF0000} \textbf{0.400}} & 0.419                              & {\textcolor{blue} {\ul 0.406}} & 0.416        & 0.417        & 0.414      & 0.425      & {\textcolor{blue} {\ul 0.406}} & 0.418 & 0.413                              & 0.416 \\
                        & 720 & 0.478        & 0.455        & {\color[HTML]{FF0000} \textbf{0.456}} & {\color[HTML]{FF0000} \textbf{0.434}} & 0.479                              & {\textcolor{blue} {\ul 0.440}} & 0.482        & 0.458        & 0.476      & 0.463      & {\textcolor{blue} {\ul 0.462}} & 0.451 & 0.473                              & 0.452 \\ \cmidrule(rl){2-16}
\multirow{-5}{*}{\rotatebox{90}{ETTm1}} & Avg & 0.405        & 0.409        & {\color[HTML]{FF0000} \textbf{0.386}} & {\color[HTML]{FF0000} \textbf{0.393}} & 0.410                              & {\textcolor{blue} {\ul 0.400}} & 0.406        & 0.410        & 0.401      & 0.414      & {\textcolor{blue} {\ul 0.394}} & 0.410 & 0.405                              & 0.412 \\ \bottomrule
\end{tabular}}
\label{tab: tab3_all}
\end{table*}
\begin{table*}[h]
\caption{Comparison between CDFM and non-stationary models with prediction lengths \(H \in
\{96, 192, 336, 720\}\) and the input length \(L = 96\).  Best results are highlighted in {\color[HTML]{FF0000} \textbf{red}} and the second {\textcolor{blue} {\ul underlined}}.}
\centering
\large
\renewcommand{\arraystretch}{0.9}
\resizebox{0.65\textwidth}{!}{
\begin{tabular}{c|c|cc|cc|cc|cc|cc}
\toprule
\multicolumn{2}{c|}{Models}      & \multicolumn{2}{c|}{\textbf{\begin{tabular}[c]{@{}c@{}}CDFM\\ (Ours)\end{tabular}}} & \multicolumn{2}{c|}{\begin{tabular}[c]{@{}c@{}}TFPS\\ (2024)\end{tabular}}     & \multicolumn{2}{c|}{\begin{tabular}[c]{@{}c@{}}HTV-Trans\\ (2024)\end{tabular}} & \multicolumn{2}{c|}{\begin{tabular}[c]{@{}c@{}}Koopa\\ (2023)\end{tabular}} & \multicolumn{2}{c}{\begin{tabular}[c]{@{}c@{}}NS-Trans\\ (2022)\end{tabular}} \\ \midrule
\multicolumn{2}{c|}{Metric}      & MSE                                      & MAE                                     & MSE                                   & MAE                                   & MSE                    & MAE                                                   & MSE                                  & MAE                                 & MSE                                   & MAE                                   \\ \midrule
                          & 96  & {\color[HTML]{FF0000} \textbf{0.374}}    & {\color[HTML]{FF0000} \textbf{0.389}}   & 0.398                                 & 0.413                                 & 0.399                  & {\textcolor{blue} {\ul 0.405}}                  & {\textcolor{blue} {\ul 0.390}}                                & {\textcolor{blue} {\ul 0.405}}                                       & 0.593                                 & 0.522                                 \\
                          & 192 & {\textcolor{blue} {\ul 0.426}}       & {\color[HTML]{FF0000} \textbf{0.419}}   & {\color[HTML]{FF0000} \textbf{0.423}} & {\textcolor{blue} {\ul 0.423}}    & 0.442                  & 0.428                                                 & 0.443                                & 0.434                               & 0.596                                 & 0.521                                 \\
                          & 336 & {\color[HTML]{FF0000} \textbf{0.465}}    & {\color[HTML]{FF0000} \textbf{0.438}}   & {\textcolor{blue} {\ul 0.484}}    & 0.461                                 & 0.490                  & {\textcolor{blue} {\ul 0.452}}                    & 0.487                                & 0.454                               & 0.756                                 & 0.594                                 \\
                          & 720 & {\color[HTML]{FF0000} \textbf{0.459}}    & {\color[HTML]{FF0000} \textbf{0.455}}   & {\textcolor{blue} {\ul 0.488}}    & 0.476                                 & 0.496                  & {\textcolor{blue} {\ul 0.473}}                    & 0.497                                & 0.480                               & 0.771                                 & 0.615                                 \\   \cmidrule(rl){2-12}
\multirow{-5}{*}{\rotatebox{90}{ETTh1}}   & Avg & {\color[HTML]{FF0000} \textbf{0.431}}    & {\color[HTML]{FF0000} \textbf{0.425}}   & {\textcolor{blue} {\ul 0.448}}    & 0.443                                 & 0.457                  & {\textcolor{blue} {\ul 0.440}}                    & 0.454                                & 0.443                               & 0.679                                 & 0.563                                 \\ \midrule
                          & 96  & {\color[HTML]{FF0000} \textbf{0.285}}    & {\color[HTML]{FF0000} \textbf{0.335}}   & 0.313                                 & 0.355                                 & 0.341                  & 0.370                                                 & {\textcolor{blue} {\ul 0.300}}   & {\textcolor{blue} {\ul 0.351}}  & 0.386                                 & 0.409                                 \\
                          & 192 & {\color[HTML]{FF0000} \textbf{0.366}}    & {\color[HTML]{FF0000} \textbf{0.385}}   & 0.405                                 & 0.410                                 & 0.435                  & 0.421                                                 & {\textcolor{blue} {\ul 0.376}}   & {\textcolor{blue} {\ul 0.400}}  & 0.506                                 & 0.468                                 \\
                          & 336 & {\color[HTML]{FF0000} \textbf{0.400}}    & {\color[HTML]{FF0000} \textbf{0.414}}   & {\textcolor{blue} {\ul 0.392}}    & {\textcolor{blue} {\ul 0.415}}    & 0.470                  & 0.458                                                 & 0.430                                & 0.441                               & 0.536                                 & 0.496                                 \\
                          & 720 & {\color[HTML]{FF0000} \textbf{0.389}}    & {\color[HTML]{FF0000} \textbf{0.416}}   & {\textcolor{blue} {\ul 0.410}}    & {\textcolor{blue} {\ul 0.433}}    & 0.478                  & 0.476                                                 & 0.443                                & 0.454                               & 0.581                                 & 0.529                                 \\  \cmidrule(rl){2-12}
\multirow{-5}{*}{\rotatebox{90}{ETTh2}}   & Avg & {\color[HTML]{FF0000} \textbf{0.360}}    & {\color[HTML]{FF0000} \textbf{0.388}}   & {\textcolor{blue} {\ul 0.380}}    & {\textcolor{blue} {\ul 0.403}}    & 0.431                  & 0.431                                                 & 0.387                                & 0.412                               & 0.502                                 & 0.475                                 \\ \midrule
                          & 96  & {\color[HTML]{FF0000} \textbf{0.324}}    & {\color[HTML]{FF0000} \textbf{0.357}}   & {\textcolor{blue} {\ul 0.327}}    & {\textcolor{blue} {\ul 0.367}}    & 0.337                  & 0.370                                                 & 0.335                                & 0.371                               & 0.405                                 & 0.408                                 \\
                          & 192 & {\color[HTML]{FF0000} \textbf{0.367}}    & {\color[HTML]{FF0000} \textbf{0.379}}   & {\textcolor{blue} {\ul 0.374}}    & 0.395                                 & 0.378                  & {\textcolor{blue} {\ul 0.388}}                    & 0.377                                & 0.392                               & 0.476                                 & 0.436                                 \\
                          & 336 & {\color[HTML]{FF0000} \textbf{0.398}}    & {\color[HTML]{FF0000} \textbf{0.400}}   & {\textcolor{blue} {\ul 0.401}}    & {\textcolor{blue} {\ul 0.408}}    & 0.418                  & 0.414                                                 & 0.411                                & 0.414                               & 0.551                                 & 0.492                                 \\
                          & 720 & {\color[HTML]{FF0000} \textbf{0.456}}    & {\color[HTML]{FF0000} \textbf{0.434}}   & 0.479                                 & 0.456                                 & 0.492                  & 0.451                                                 & {\textcolor{blue} {\ul 0.473}}   & {\textcolor{blue} {\ul 0.447}}  & 0.664                                 & 0.535                                 \\  \cmidrule(rl){2-12}
\multirow{-5}{*}{\rotatebox{90}{ETTm1}}   & Avg & {\color[HTML]{FF0000} \textbf{0.386}}    & {\color[HTML]{FF0000} \textbf{0.393}}   & {\textcolor{blue} {\ul 0.395}}    & 0.407                                 & 0.406                  & {\textcolor{blue} {\ul 0.406}}                                                 & 0.399                                & {\textcolor{blue} {\ul 0.406}}  & 0.524                                 & 0.468                                 \\ \midrule
                          & 96  & 0.184       & 0.268      & {\color[HTML]{FF0000} \textbf{0.170}} & {\color[HTML]{FF0000} \textbf{0.255}} & 0.184                  & 0.266                                                 & {\textcolor{blue} {\ul 0.182}}                                & {\textcolor{blue} {\ul 0.265}}                               & 0.217                                 & 0.300                                 \\
                          & 192 & {\textcolor{blue} {\ul 0.244}}       & {\textcolor{blue} {\ul 0.303}}      & {\color[HTML]{FF0000} \textbf{0.235}} & {\color[HTML]{FF0000} \textbf{0.296}} & 0.249                  & 0.307                                                 & 0.246                                & 0.305                               & 0.433                                 & 0.396                                 \\
                          & 336 & 0.302       & {\textcolor{blue} {\ul 0.341}}                                 & {\color[HTML]{FF0000} \textbf{0.297}} & {\color[HTML]{FF0000} \textbf{0.335}} & 0.308                  & 0.343                                                 & {\textcolor{blue} {\ul 0.301}}                                & {\textcolor{blue} {\ul 0.341}}  & 0.892                                 & 0.623                                 \\
                          & 720 & {\textcolor{blue} {\ul 0.404}}    & {\color[HTML]{FF0000} \textbf{0.396}}   &  {\color[HTML]{FF0000} \textbf{0.401}}    & {\textcolor{blue} {\ul 0.397}}    & 0.411                  & 0.401                                                 & 0.408                                & 0.404                               & 0.927                                 & 0.618                                 \\  \cmidrule(rl){2-12}
\multirow{-5}{*}{\rotatebox{90}{ETTm2}}   & Avg & {\textcolor{blue} {\ul 0.284}}       & {\textcolor{blue} {\ul 0.327}}      & {\color[HTML]{FF0000} \textbf{0.276}} & {\color[HTML]{FF0000} \textbf{0.321}} & 0.288                  & 0.329                                                 & {\textcolor{blue} {\ul 0.284}}                                 & 0.329                               & 0.617                                 & 0.484                                 \\ \midrule
                          & 96  & 0.176                                    & 0.231                                   & {\color[HTML]{FF0000} \textbf{0.154}} & {\color[HTML]{FF0000} \textbf{0.202}} & 0.174                  & 0.223                                                 & {\textcolor{blue} {\ul 0.158}}   & {\textcolor{blue} {\ul 0.204}}  & 0.180                                 & 0.227                                 \\
                          & 192 & 0.213                                    & 0.263                                   & {\color[HTML]{FF0000} \textbf{0.205}} & {\color[HTML]{FF0000} \textbf{0.249}} & 0.222                  & {\textcolor{blue} {\ul 0.261}}                                                & {\textcolor{blue} {\ul 0.207}}   & {\color[HTML]{FF0000} \textbf{0.249}}  & 0.252                                 & 0.293                                 \\
                          & 336 & {\color[HTML]{FF0000} \textbf{0.257}}    & 0.300                                   & {\textcolor{blue} {\ul 0.262}}    & {\color[HTML]{FF0000} \textbf{0.289}} & 0.280                  & 0.301                                                 & 0.266                                & {\textcolor{blue} {\ul 0.291}}  & 0.320                                 & 0.339                                 \\
                          & 720 & {\color[HTML]{FF0000} \textbf{0.317}}    & {\color[HTML]{FF0000} \textbf{0.339}}     & {\textcolor{blue} {\ul 0.344}}    & {\textcolor{blue} {\ul 0.342}}  & 0.357                  & 0.349                                                 & 0.347                                & 0.344                               & 0.377                                 & 0.370                                 \\  \cmidrule(rl){2-12}
\multirow{-5}{*}{\rotatebox{90}{Weather}} & Avg & {\color[HTML]{FF0000} \textbf{0.241}}    & 0.283                                   & {\color[HTML]{FF0000} \textbf{0.241}} & {\color[HTML]{FF0000} \textbf{0.271}} & 0.258                  & 0.284                                                 & 0.245                                & {\textcolor{blue} {\ul 0.272}}  & 0.282                                 & 0.307             \\ \midrule

                              & 96  & 0.191 & 0.272                              & {\textcolor{blue} {\ul  0.149}}    & {\color[HTML]{FF0000} \textbf{0.236}} & 0.193 & 0.295 & {\color[HTML]{FF0000} \textbf{0.147}} & {\textcolor{blue} {\ul  0.247}}    & 0.169                              & 0.273 \\
                              & 192 & 0.191 & 0.278                              & {\color[HTML]{FF0000} \textbf{0.162}} & {\color[HTML]{FF0000} \textbf{0.253}} & 0.197 & 0.295 & {\textcolor{blue} {\ul  0.181}}    & {\textcolor{blue} {\ul  0.276}}    & 0.182                              & 0.286 \\
                              & 336 & 0.205 & {\textcolor{blue} {\ul  0.294}} & {\textcolor{blue} {\ul  0.200}}    & 0.310                                 & 0.213 & 0.310 & {\color[HTML]{FF0000} \textbf{0.199}} & {\color[HTML]{FF0000} \textbf{0.292}} & {\textcolor{blue} {\ul  0.200}} & 0.304 \\
                              & 720 & 0.242 & 0.327                              & {\color[HTML]{FF0000} \textbf{0.220}} & {\textcolor{blue} {\ul  0.320}}    & 0.251 & 0.341 & 0.228                                 & {\color[HTML]{FF0000} \textbf{0.316}} & {\textcolor{blue} {\ul  0.222}} & 0.321 \\
\multirow{-5}{*}{\rotatebox{90}{Electricity}} & Avg & 0.207 & 0.293                              & {\color[HTML]{FF0000} \textbf{0.183}} & {\color[HTML]{FF0000} \textbf{0.280}} & 0.214 & 0.310 & 0.189                                 & 0.283                                 & 0.193                              & 0.296 \\ \midrule
\rowcolor{pink!50}
\multicolumn{2}{c|}{$1^\text{st}$ Count}           & {\color[HTML]{FF0000} \textbf{17}}    & {\color[HTML]{FF0000} \textbf{17}}                                 & {\textcolor{blue} {\ul 12}}        & {\textcolor{blue} {\ul 11}}        & 0                                   & 0                                   & 2                                     &3                                     & 0                                                    & 0                     \\ 

\bottomrule                   
\end{tabular}}

\label{tab: tab4_all}
\end{table*}

\begin{table*}[t]
\caption{Performance promotion by applying the proposed framework to other models with \(L=96\). IMP. represents the relative MSE reduction.}
\centering
\large
\renewcommand{\arraystretch}{0.9}
\resizebox{0.9\textwidth}{!}{
\begin{tabular}{c|c|cc|cc|c|cc|cc|c|cc|cc|c}
\toprule
\multicolumn{2}{c|}{Models}          & \multicolumn{2}{c|}{DLinear} & \multicolumn{2}{c|}{+CDFM}                                                                                              & IMP.   & \multicolumn{2}{c|}{TimeMixer} & \multicolumn{2}{c|}{+CDFM}                                                      & IMP.  & \multicolumn{2}{c|}{iTransformer}              & \multicolumn{2}{c|}{+CDFM}                                                      & IMP.   \\ \midrule
\multicolumn{2}{c}{Metric}          & MSE          & MAE          & MSE                                                       & MAE                                                       & MSE    & MSE           & MAE           & MSE                                   & MAE                                   & MSE   & MSE   & MAE                                   & MSE                                   & MAE                                   & MSE    \\ \midrule
                              & 96  & 0.315        & 0.374        & {\color[HTML]{FF0000} \textbf{0.285}}                     & {\color[HTML]{FF0000} \textbf{0.335}}                     & 9.5\%                  & 0.295                                  & 0.346                                 & {\color[HTML]{FF0000} \textbf{0.286}} & {\color[HTML]{FF0000} \textbf{0.339}} & 3.1\%                  & 0.301                   & 0.350                                                   & {\color[HTML]{FF0000} \textbf{0.299}} & {\color[HTML]{FF0000} \textbf{0.350}} & 0.7\%                  \\
                              & 192 & 0.432        & 0.447        & {\color[HTML]{FF0000} \textbf{0.366}}                     & {\color[HTML]{FF0000} \textbf{0.385}}                     & 15.3\%                 & 0.376                                  & 0.395                                 & {\color[HTML]{FF0000} \textbf{0.368}} & {\color[HTML]{FF0000} \textbf{0.389}} & 2.1\%                  & 0.380                   & {\color[HTML]{FF0000} \textbf{0.399}}                   & {\color[HTML]{FF0000} \textbf{0.379}} & 0.400                                 & 0.3\%                  \\
                              & 336 & 0.486        & 0.481        & \multicolumn{1}{l}{{\color[HTML]{FF0000} \textbf{0.400}}} & {\color[HTML]{FF0000} \textbf{0.414}}                     & 17.7\%                 & 0.421                                  & 0.435                                 & {\color[HTML]{FF0000} \textbf{0.412}} & {\color[HTML]{FF0000} \textbf{0.425}} & 2.1\%                  & 0.424                   & 0.432                                                   & {\color[HTML]{FF0000} \textbf{0.420}} & {\color[HTML]{FF0000} \textbf{0.430}} & 0.9\%                  \\
\multirow{-4}{*}{\rotatebox{90}{ETTh2}}       & 720 & 0.732        & 0.614        & {\color[HTML]{FF0000} \textbf{0.389}}                     & {\color[HTML]{FF0000} \textbf{0.416}}                     & 46.9\%                 & 0.445                                  & 0.458                                 & {\color[HTML]{FF0000} \textbf{0.428}} & {\color[HTML]{FF0000} \textbf{0.44}}  & 3.8\%                  & 0.430                   & 0.447                                                   & {\color[HTML]{FF0000} \textbf{0.409}} & {\color[HTML]{FF0000} \textbf{0.430}} & 4.9\%                  \\ \midrule
                              & 96  & 0.197        & 0.257        & {\color[HTML]{FF0000} \textbf{0.176}}                     & {\color[HTML]{FF0000} \textbf{0.231}}                     & 10.7\%                 & 0.162                                  & 0.209                                 & {\color[HTML]{FF0000} \textbf{0.160}} & {\color[HTML]{FF0000} \textbf{0.208}} & 1.1\%                  & 0.176                   & 0.216                                                   & {\color[HTML]{FF0000} \textbf{0.165}} & {\color[HTML]{FF0000} \textbf{0.206}} & 6.2\%                  \\
                              & 192 & 0.237        & 0.294        & {\color[HTML]{FF0000} \textbf{0.213}}                     & {\color[HTML]{FF0000} \textbf{0.263}}                     & 10.1\%                 & 0.207                                  & 0.251                                 & {\color[HTML]{FF0000} \textbf{0.201}} & {\color[HTML]{FF0000} \textbf{0.251}} & 3.0\%                  & 0.225                   & 0.257                                                   & {\color[HTML]{FF0000} \textbf{0.213}} & {\color[HTML]{FF0000} \textbf{0.251}} & 5.3\%                  \\
                              & 336 & 0.283        & 0.332        & {\color[HTML]{FF0000} \textbf{0.257}}                     &{\color[HTML]{FF0000} \textbf{0.300}} & 9.2\%                  & 0.264                                  & 0.294                                 & {\color[HTML]{FF0000} \textbf{0.251}} & {\color[HTML]{FF0000} \textbf{0.288}} & 5.0\%                  & 0.281                   & 0.299                                                   & {\color[HTML]{FF0000} \textbf{0.263}} & {\color[HTML]{FF0000} \textbf{0.291}} & 6.4\%                  \\
\multirow{-4}{*}{\rotatebox{90}{Weather}}     & 720 & 0.347        & 0.382        & {\color[HTML]{FF0000} \textbf{0.317}}                     & {\color[HTML]{FF0000} \textbf{0.339}}                     & 8.6\%                  & 0.345                                  & 0.348                                 & {\color[HTML]{FF0000} \textbf{0.323}} & {\color[HTML]{FF0000} \textbf{0.341}} & 6.4\%                  & 0.358                   & 0.350                                                   & {\color[HTML]{FF0000} \textbf{0.331}} & {\color[HTML]{FF0000} \textbf{0.337}} & 7.5\%                  \\ \midrule
                              & 96  & 0.195        & 0.277        & {\color[HTML]{FF0000} \textbf{0.191}}                     & {\color[HTML]{FF0000} \textbf{0.272}}                     & 2.1\%                  & 0.153                                  & 0.247                                 & {\color[HTML]{FF0000} \textbf{0.152}} & {\color[HTML]{FF0000} \textbf{0.245}} & 0.7\%                  & 0.151                   & 0.241                                                   & {\color[HTML]{FF0000} \textbf{0.143}} & {\color[HTML]{FF0000} \textbf{0.236}} & 5.3\%                  \\
                              & 192 & 0.194        & 0.281        & {\color[HTML]{FF0000} \textbf{0.191}}                     & {\color[HTML]{FF0000} \textbf{0.278}}                     & 1.5\%                  & 0.166                                  & 0.258                                 & {\color[HTML]{FF0000} \textbf{0.164}} & {\color[HTML]{FF0000} \textbf{0.256}} & 1.2\%                  & 0.167                   & 0.258                                                   & {\color[HTML]{FF0000} \textbf{0.159}} & {\color[HTML]{FF0000} \textbf{0.252}} & 4.8\%                  \\
                              & 336 & 0.207        & 0.296        & {\color[HTML]{FF0000} \textbf{0.205}}                     & {\color[HTML]{FF0000} \textbf{0.294}}                     & 1.0\%                  & 0.186                                  & 0.277                                 & {\color[HTML]{FF0000} \textbf{0.181}} & {\color[HTML]{FF0000} \textbf{0.275}} & 2.7\%                  & 0.179                   & 0.271                                                   & {\color[HTML]{FF0000} \textbf{0.172}} & {\color[HTML]{FF0000} \textbf{0.265}} & 3.9\%                  \\
\multirow{-4}{*}{\rotatebox{90}{Electricity}} & 720 & 0.243        & 0.330        & {\color[HTML]{FF0000} \textbf{0.242}}                     & {\color[HTML]{FF0000} \textbf{0.327}}                     & 0.4\%                  & 0.225                                  & 0.312                                 & {\color[HTML]{FF0000} \textbf{0.218}} & {\color[HTML]{FF0000} \textbf{0.309}} & 3.1\%                  & 0.229                   & 0.319                                                   & {\color[HTML]{FF0000} \textbf{0.198}} & {\color[HTML]{FF0000} \textbf{0.291}} & 13.5\%                
\\ \bottomrule
\end{tabular}}
\label{tab: tab5_all}
\vspace{2mm}
\end{table*}

\end{document}